\documentclass[lettersize,journal]{IEEEtran}
\usepackage{amsmath,amsfonts}
\usepackage{algorithmic}
\usepackage{algorithm}
\usepackage{array}
\usepackage[caption=false,font=normalsize,labelfont=sf,textfont=sf]{subfig}
\usepackage{textcomp}
\usepackage{stfloats}
\usepackage{url}
\usepackage{verbatim}
\usepackage{graphicx}
\usepackage{cite}
\usepackage{multirow}
\usepackage{booktabs}
\usepackage{hyperref}
\hypersetup{colorlinks=true, urlcolor=magenta}
\usepackage[table]{xcolor}
\definecolor{firstcolor}{HTML}{BDE6CD}
\definecolor{secondcolor}{HTML}{E2EEBC}
\definecolor{thirdcolor}{HTML}{FFF8C5}
\newcommand{\fst}[1]{\cellcolor{firstcolor}#1}
\newcommand{\snd}[1]{\cellcolor{secondcolor}#1}
\newcommand{\trd}[1]{\cellcolor{thirdcolor}#1}

\begin{document}

\title{Geometry Reinforced Efficient Attention Tuning Equipped with Normals for Robust Stereo Matching}

\author{Jiahao Li, Xinhong Chen, Zhengmin JIANG, Cheng Huang, Yung-Hui Li~\IEEEmembership{Member,~IEEE,}, Jianping Wang,~\IEEEmembership{Fellow,~IEEE,}
\thanks{Jiahao Li, Xinhong Chen, Zhengmin JIANG, and Jianping Wang are with the Department of Computer Science, City University of Hong Kong, Hong Kong, China (E-mail: jiahali2-c@my.cityu.edu.hk, \{xinhong.chen, zhengmin.jiang, jianwang\}@cityu.edu.hk).}
\thanks{Cheng Huang is with the Department of Computer Science, Southern Methodist University, Dallas, TX, 75205, USA (E-mail: chenghuang@smu.edu).}
\thanks{Yung-Hui Li is with the Hon Hai Research Institute (E-mail: yunghui.li@foxconn.com). He is a member of IEEE.}
\thanks{Corresponding Author: Jianping Wang, she is a fellow of IEEE and a fellow of AAIA.}
}




\maketitle

\begin{abstract}
Despite remarkable advances in image-driven stereo matching over the past decade, Synthetic-to-Realistic Zero-Shot (Syn-to-Real) generalization remains an open challenge. This suboptimal generalization performance mainly stems from cross-domain shifts and ill-posed ambiguities inherent in image textures, particularly in occluded, textureless, repetitive, and non-Lambertian (specular/transparent) regions. To tackle this, we propose GREATEN, an advanced framework that builds upon our prior GREAT architecture to achieve robust zero-shot capabilities. To compensate for the inherent unreliability of image textures, GREATEN introduces a paradigm shift by incorporating surface normals as domain-invariant, object-intrinsic, and discriminative geometric cues. The proposed framework features three key innovations. First, a Gated Contextual-Geometric Fusion (GCGF) module adaptively suppresses unreliable contextual cues in image features and fuses the filtered image features with normal-driven geometric features to construct domain-invariant and discriminative contextual-geometric representations. Second, a Specular-Transparent Augmentation (STA) strategy improves the robustness of GCGF against misleading visual cues in non-Lambertian regions. Third, sparse attention designs preserve the fine-grained global feature extraction capability of the original GREAT-Stereo for handling occlusion and texture-related ambiguities while substantially reducing computational overhead, including Sparse Spatial (SSA), Sparse Dual-Matching (SDMA), and Simple Volume (SVA) attentions. Trained exclusively on synthetic data such as SceneFlow, GREATEN-IGEV achieves outstanding Syn-to-Real performance. Specifically, it reduces errors by 30\% on ETH3D, 8.5\% on the non-Lambertian Booster, and 14.1\% on KITTI-2015, compared to FoundationStereo, Monster-Stereo, and DEFOM-Stereo, respectively. In addition, GREATEN-IGEV runs 19.2\% faster than its predecessor, GREAT-IGEV, and supports high-resolution (3K) inference on Middlebury with disparity ranges up to 768. Code is available at \url{https://github.com/JarvisLee0423/GREAT-Stereo} and \url{https://github.com/JarvisLee0423/GREATEN-Stereo}.

\end{abstract}


\section{Introduction}
\label{sec:intro}

\IEEEPARstart{S}{tereo} matching recovers dense 3D representations from rectified image pairs by regressing the displacements between corresponding pixel locations in the left and right views, referred to as disparity \cite{sgbm}. It is a fundamental component in many real-world applications, including 3D reconstruction, robotic perception, and autonomous driving.

\begin{figure}[tp]
    \centering
    \includegraphics[width=0.99\linewidth]{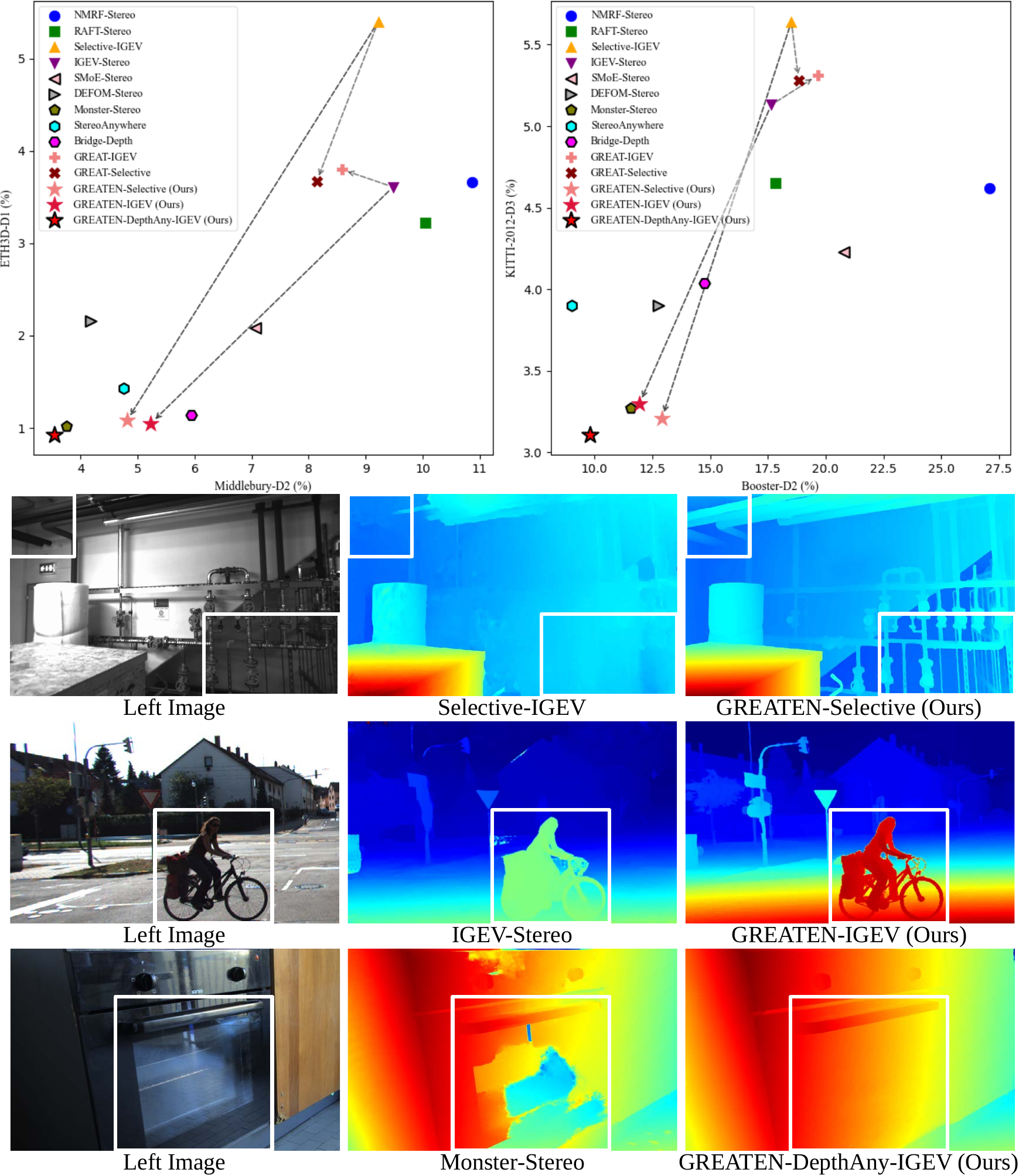}
    \caption{\textbf{Row 1}: Comparison of Syn-to-Real generalization on ETH3D \cite{eth3d}, Middlebury \cite{middlebury}, KITTI-2012 \cite{kitti2012}, and Booster \cite{booster}, where the lower metrics indicate better performance (Thick boundary methods use Vision-Foundation-Model \cite{depthanythingv2}). \textbf{Row 2}: Visual comparison with Selective-IGEV \cite{selectivestereo} on ETH3D. \textbf{Row 3}: Visual comparison with IGEV-Stereo \cite{igevstereo} on KITTI-2015 \cite{kitti2015}. \textbf{Row 4}: Visual comparison with Monster-Stereo \cite{monsterstereo} on Booster. Our models outperform previous methods, especially in ill-posed regions (white boxes).}
    \label{fig:performance_comparison}
\end{figure}


Recent stereo matching algorithms can be primarily categorized into two distinct paradigms. Aggregation-based approaches \cite{gcnet, psmnet, gwcnet, acvnet, fastacvnet, dcanet, adnet, dualnet} leverage 3D convolutional neural networks (CNNs) to regularize cost volumes before disparity regression, but suffer from substantial computational overhead. On the other hand, iterative schemes \cite{raftstereo, crestereo, spstereo, pcvstereo, igevstereo, selectivestereo, greatstereo} replace computationally intensive 3D convolutions with Convolutional Gated Recurrent Units (ConvGRUs) to progressively refine disparities via cost-volume look-up. Despite these efforts, existing methods are typically trained on synthetic data (e.g., SceneFlow \cite{sceneflow}) due to the scarcity of high-quality real-world stereo datasets. Hence, most of them encounter challenges in bridging the synthetic and realistic domains, as reflected by the suboptimal Synthetic-to-Realistic (Syn-to-Real) generalization performance shown in Fig. \ref{fig:performance_comparison}.


\begin{figure*}[tp]
    \centering
    \includegraphics[width=0.97\linewidth]{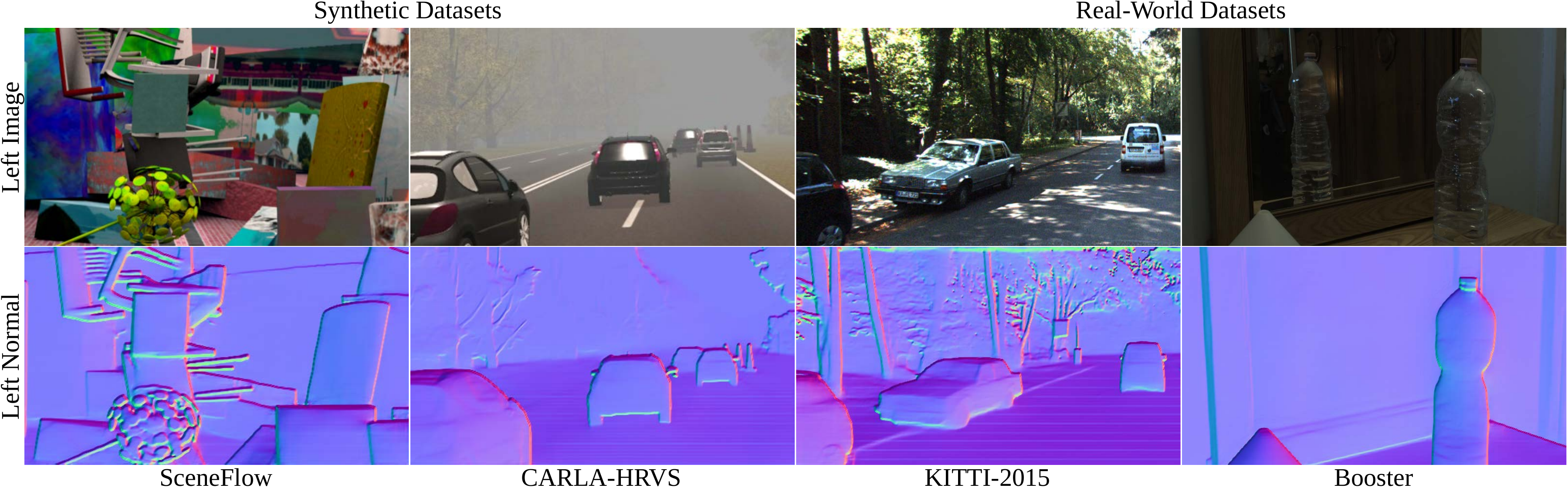}
    \caption{Comparison of domain shifts between images and surface normals across synthetic-to-realistic datasets. Surface normals exhibit domain invariance, preserving consistent representations across synthetic and real-world domains. In contrast, images demonstrate substantial domain gaps, even under similar autonomous driving scenarios (CARLA-HRVS \cite{hrvs} vs. KITTI-2015 \cite{kitti2015}).}
    \label{fig:comparison_of_images_and_normals}
\end{figure*}

To address this challenge, existing methods predominantly focus on mitigating the cross-domain shifts within image textures, such as variations in illumination, noise patterns, and overall appearance, as illustrated in Fig. \ref{fig:comparison_of_images_and_normals} ("CARLA-HRVS" vs "KITTI-2015"). Specifically, domain transfer and adaptation approaches \cite{graftnet, sdanet, hvtnet, maskednet} employ Generative Adversarial Networks (GANs) \cite{gans} or Masked Autoencoders (MAEs) \cite{mae} to learn domain-adaptive representations. Meanwhile, domain generalization methods \cite{dsmnet, formernet, defomstereo, monsterstereo, foundationstereo} seek cross-domain robustness through domain-invariant normalization or Vision Foundation Models (VFMs) \cite{dinov2, depthanythingv2}. However, all these methods fail to effectively handle the inherently ambiguous and misleading textures within ill-posed regions, such as occluded, textureless, repetitive, and non-Lambertian (specular/transparent) areas. For instance, reflections in non-Lambertian regions frequently generate superimposed textures that combine the intrinsic appearance of the target object with imaginary foreground objects, as exemplified by the "Booster" case in Fig. \ref{fig:comparison_of_images_and_normals}. Although GANs and VFMs demonstrate strong domain transfer and zero-shot generalization capabilities, the inherently ambiguous textures in these ill-posed areas continue to mislead them \cite{stereoanywhere}, resulting in generalization failures, as illustrated in Fig. \ref{fig:performance_comparison}, Row 4.


To this end, we argue that robust Syn-to-Real generalization requires an information source that is both domain-invariant and sufficiently discriminative to compensate for the inherent limitations of image textures. As illustrated in Fig. \ref{fig:comparison_of_images_and_normals}, surface normals provide precisely such a cue. Unlike textures, surface normals describe unit vectors perpendicular to object surfaces, thereby encoding object-intrinsic geometric properties that remain strictly invariant across domains. Furthermore, they offer critical discriminative information in ill-posed regions where visual appearance is inherently ambiguous (e.g., the "Booster" case in Fig. \ref{fig:comparison_of_images_and_normals}). However, Fig. \ref{fig:comparison_of_images_and_normals} also reveals that surface normals alone are less informative than image textures, demonstrating relatively "flat" when searching for corresponding pixel locations. Therefore, to preserve strong in-domain performance while improving cross-domain generalization, stereo matching should effectively integrate the complementary strengths of image textures and surface normals.


In this paper, we propose a universal framework termed \textbf{G}eometry \textbf{R}einforced \textbf{E}fficient \textbf{A}ttention \textbf{T}uning \textbf{E}quipped with \textbf{N}ormals (\textbf{GREATEN}). Extending our preliminary work, GREAT \cite{greatstereo}, GREATEN shifts focus to robust zero-shot generalization. It extracts image contextual and normal geometric features via dual encoders, integrating them through a \textbf{Gated Contextual-Geometric Fusion} (GCGF) module. GCGF adaptively masks ambiguous visual cues and fuses the refined visual features with normal features to yield domain-invariant representations. This is further enhanced by a \textbf{Specular-Transparent Augmentation} (STA) strategy that perturbs training textures to force better filtering and fusion. Finally, to resolve the computational overhead of GREAT \cite{greatstereo} while inheriting its global context modeling for ambiguities, we evolve its dense attentions into efficient sparse alternatives: \textbf{Sparse Spatial Attention} (SSA), \textbf{Sparse Dual-Matching Attention} (SDMA), and \textbf{Simple Volume Attention} (SVA).

The proposed GREATEN framework achieves outstanding Syn-to-Real generalization across five widely adopted realistic benchmarks: ETH3D \cite{eth3d}, Middlebury \cite{middlebury}, KITTI-2012 \cite{kitti2012}, KITTI-2015 \cite{kitti2015}, and Booster \cite{booster}. When integrated with IGEV-Stereo \cite{igevstereo} and trained exclusively on the synthetic SceneFlow \cite{sceneflow} dataset, our VFM-Free GREATEN-IGEV achieves EPE of 2.85 on the non-Lambertian Booster \cite{booster}. Remarkably, it outperforms several VFM-Enhanced methods, reducing the EPE of Bridge-Depth \cite{bridgedepth} by 30\%. Furthermore, when trained on our Syn-to-Real mixed datasets, both GREATEN-IGEV and its VFM-Enhanced GREATEN-DepthAny-IGEV achieve state-of-the-art generalization performance across all five real-world benchmarks. In addition, GREATEN also exhibits outstanding in-domain robustness, ranking first on both the KITTI-2015 \cite{kitti2015} and ETH3D \cite{eth3d} leaderboards under the Robust Vision Challenge (RVC) settings. Moreover, benefiting from the proposed sparse attention modules, GREATEN reduces inference time by 18.5\% on SceneFlow \cite{sceneflow} and 19.2\% on Middlebury \cite{middlebury} compared to the original GREAT \cite{greatstereo} framework, while also enabling full-resolution inference on Middlebury \cite{middlebury} with disparity ranges up to 768.

A preliminary version of this work was presented at ICCV 2025 as GREAT \cite{greatstereo}. In this journal extension, we significantly advance the framework to GREATEN to tackle the fundamentally more challenging problem of Syn-to-Real zero-shot generalization. Compared to the conference version, our new contributions are summarized as follows:

\begin{itemize}
    \item We introduce the Gated Contextual-Geometric Fusion (GCGF) module that effectively fuses stereo-image and surface-normal features to mitigate cross-domain discrepancies and the ill-posed ambiguities inherent in image textures, thereby enhancing Synthetic-to-Realistic (Syn-to-Real) generalization.
    \item We design the Specular-Transparent Augmentation (STA) strategy to intentionally disturb the texture consistency of training images, forcing the GCGF module to better filter ambiguous image textures and improve fusion reliability.
    \item We develop sparse attention alternatives to preserve the global feature extraction capability of our predecessor GREAT-Stereo \cite{greatstereo} for handling ambiguities in occluded and texture-related ill-posed regions, while significantly reducing computational cost, termed as Sparse Spatial Attention (SSA), Sparse Dual-Matching Attention (SDMA), and Simple Volume Attention (SVA).
    \item Trained solely on synthetic data, our GREATEN-Stereo outperforms existing published stereo-matching methods in Synthetic-to-Realistic generalization across five major real-world benchmarks: ETH3D \cite{eth3d}, Middlebury \cite{middlebury}, KITTI-2012 \cite{kitti2012}, KITTI-2015 \cite{kitti2015}, and Booster \cite{booster}.
\end{itemize}

\section{Related Work}
\label{sec:related_work}

\subsection{Aggregation-based \& Iterative Approaches}
\label{sec:related_workd_A}

Stereo matching has developed rapidly in recent years. Most methods follow a standard pipeline that constructs a cost volume by measuring correlations between candidate corresponding pixels, followed by disparity regression. Aggregation-based methods \cite{psmnet, rsnet, coatrsnet, gwcnet, gcnet, pcwnet, smoestereo, acvnet, fastacvnet, aanet, ganet, dispnet} typically rely on 3D CNNs to regularize the cost volume for disparity estimation, including DispNet \cite{dispnet}, GCNet \cite{gcnet}, PSMNet \cite{psmnet}, and ACVNet \cite{acvnet}. Despite their remarkable performance, the substantial computational demands of 3D CNNs limit their efficiency, especially during high-resolution inference. To address this issue, cascaded methods \cite{cascademvsnet, cfnet} have been introduced to reduce computational cost via a coarse-to-fine strategy. However, this design is prone to error propagation from coarse disparity maps. Inspired by RAFT \cite{raft}, iterative methods \cite{raftstereo, crestereo, dlnrstereo, pcvstereo, igevstereo, selectivestereo, igev++} have recently emerged as an attractive alternative, offering a favorable balance between efficiency and accuracy. RAFT-Stereo \cite{raftstereo} first established this paradigm by progressively refining disparities through cost-volume look-up. Among these iterative approaches, IGEV-Stereo \cite{igevstereo} marks an important advance by combining iterative refinement with cost volume aggregation using lightweight 3D CNNs to capture non-local image context. Building on these developments, our GREATEN framework follows iterative schemes to recurrently update high-resolution disparity maps with ConvGRUs.

\subsection{Synthetic and Realistic Domain Shifts in Stereo Matching}
\label{sec:related_workd_B}

To alleviate the cross-domain generalization challenge, several lines of research have been explored \cite{graftnet, sdanet, hvtnet, maskednet, sssmnet, activenet, pvnet, panet, dualnet}. Domain transfer and adaptation methods aim to reduce discrepancies across domains by learning domain-adaptive representations. Among them, SDANet \cite{sdanet} employs Generative Adversarial Networks (GANs) \cite{gans} to transfer image representations from the synthetic domain to the realistic domain. MaskedNet \cite{maskednet} utilizes Masked Autoencoders \cite{mae} to regularize the learning of the features in the backbone and produce domain-adaptive representations. Similarly, DualNet \cite{dualnet} uses a well-trained teacher model to guide feature learning. More recently, domain generalization methods \cite{dsmnet, formernet, defomstereo, monsterstereo, foundationstereo} have attracted increasing attention. In particular, FoundationStereo \cite{foundationstereo} and Monster-Stereo \cite{monsterstereo} leverage strong cross-domain priors from Vision Foundation Models (VFMs) \cite{depthanythingv2} to improve Syn-to-Real zero-shot performance in stereo matching. Nevertheless, the ambiguities inherent in ill-posed textures can still mislead VFMs and result in generalization failures (Fig. \ref{fig:performance_comparison}, Row 4). To further enhance Syn-to-Real generalization, our proposed GREATEN framework incorporates domain-invariant geometric cues derived from surface normals, enabling strong cross-domain zero-shot performance even without relying on VFM guidance.

\subsection{Ill-Posed Ambiguities in Stereo Matching}
\label{sec:related_workd_C}

Besides texture discrepancies between synthetic and realistic domains, ambiguities in ill-posed regions, including occluded, textureless, repetitive, and non-Lambertian (specular/transparent) areas, are another critical factor of Syn-to-Real generalization failure. To address them, uncertainty-guided methods \cite{crestereo++, ercnet, goatstereo, uenet, dlnrstereo} have been proposed to improve occlusion handling. CREStereo++ \cite{crestereo++}, GOAT-Stereo \cite{goatstereo}, and ERCNet \cite{ercnet} predict uncertainty maps as error indicators to detect occlusions and reduce ill-posed ambiguities. Inspired by GMA-Flow \cite{gmaflow}, GREAT-Stereo \cite{greatstereo} addresses multiple types of ill-posedness, including occluded, textureless, and repetitive regions, through global feature extraction. However, these methods remain ineffective in handling non-Lambertian ambiguities and generally fail in Syn-to-Real generalization. StereoAnywhere \cite{stereoanywhere} is among the few stereo matching methods that can address non-Lambertian ambiguities under Syn-to-Real generalization. Specifically, it constructs an additional cost volume from surface normals to reduce non-Lambertian noise in the texture-based cost volume. To improve robustness across all types of ill-posed issues in Syn-to-Real generalization, our GREATEN-Stereo effectively fuses features derived from image textures and surface normals to form domain-invariant and discriminative representations. Moreover, by inheriting the global feature extraction capability of GREAT-Stereo \cite{greatstereo} via sparse attentions, GREATEN-Stereo achieves excellent Syn-to-Real generalization in ill-posed areas while substantially reducing computational overhead.

\begin{figure*}[tp]
    \centering
    \includegraphics[width=0.97\linewidth]{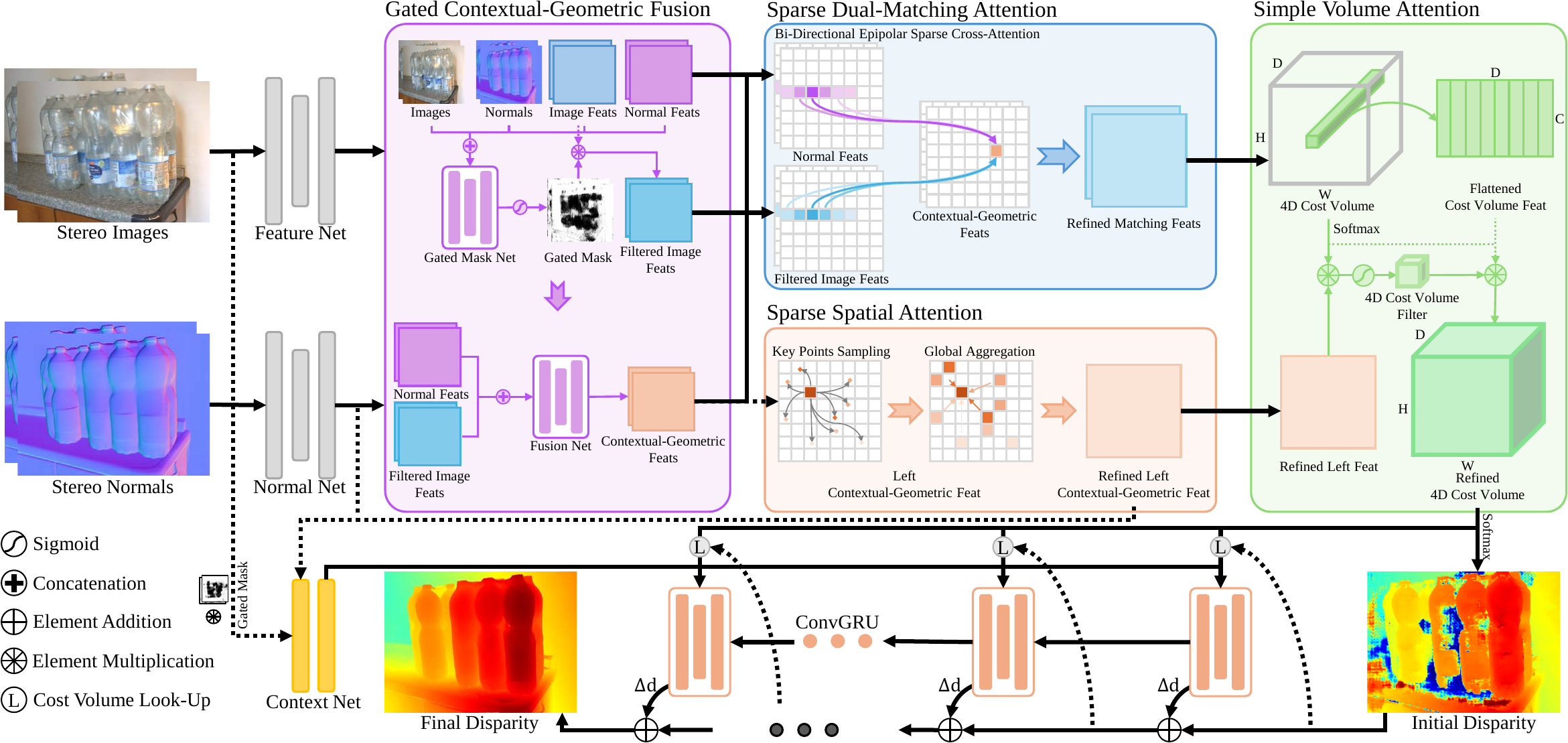}
    \caption{Overview of the proposed GREATEN framework (GREATEN-IGEV version). GREATEN-IGEV initially employs a Gated Contextual-Geometric Fusion (GCGF) module to fuse image and surface normal features ("Image Feats" \& "Normal Feats"). Subsequently, Sparse Dual-Matching Attention (SDMA) encodes global context and geometric structures within these features for cost volume construction. Finally, Sparse Spatial Attention (SSA) combines with Simple Volume Attention (SVA) to refine the cost volume and regress the initial disparity map, which undergoes iterative refinement via ConvGRU.}
    \label{fig:main_architecture}
\end{figure*}

\section{Method}
\label{sec:method}

In this section, we introduce GREATEN-Stereo, a universal framework designed to enhance the synthetic-to-realistic (Syn-to-Real) generalization in both regular and ill-posed regions for various iterative stereo-matching methods. We present GREATEN-IGEV (Fig. \ref{fig:main_architecture}) as a representative example and detail its key components.

\subsection{Framework Outline}
\label{sec:method_outline}

Given stereo images $ \mathbf{I}_{l(r)} $ and surface normals $ \mathbf{N}_{l(r)} $ in $ \mathbb{R}^{3 \times H \times W} $, where $ l $ and $ r $ denote left and right, GREATEN-IGEV first employs MobileNetV2 \cite{mobilenetv2} and Simple U-Net to extract multi-scale image and surface normal features ($ \mathbf{f}_{I_{l(r)}}^{i} $ and $ \mathbf{f}_{N_{l(r)}}^{i} \in \mathbb{R}^{C_i \times \frac{H}{i} \times \frac{W}{i}} $, where $ i \in \{4, 8, 16, 32\} $ and $ C_i $ represents feature channels). Exploiting the complementary strengths of informative image textures and domain-invariant surface normals, \textbf{Gated Contextual-Geometric Fusion} (GCGF, Sec. \ref{sec:method_gcgf}) generates discriminative and domain-invariant contextual-geometric features $ \mathbf{f}_{{l(r)}}^{fuse, i} \in \mathbb{R}^{C_i \times \frac{H}{i} \times \frac{W}{i}} $ for robust Syn-to-Real generalization. To further improve fusion robustness within the GCGF module, \textbf{Specular-Transparent Augmentation} (STA, Sec. \ref{sec:method_sta}) strategy intentionally disrupts texture consistency in the training images. Subsequently, \textbf{Sparse Dual-Matching Attention} (SDMA, Sec. \ref{sec:method_sparse}) utilizes the largest-scale features $ \mathbf{f}_{I_{l(r)}}^{4} $ and $ \mathbf{f}_{N_{l(r)}}^{4} $ to capture fine-grained global context and geometric structures along epipolar lines, facilitating the construction of a combined cost volume $ \mathbf{C}_{comb} $. In parallel, \textbf{Sparse Spatial Attention} (SSA, Sec. \ref{sec:method_sparse}) aggregates global contextual-geometric details from $ \mathbf{f}_{l}^{fuse, i} $ and integrates them with \textbf{Simple Volume Attention} (SVA, Sec. \ref{sec:method_sparse}) to produce a robust cost volume for disparity initialization and refinement.

\subsection{Gated Contextual-Geometric Fusion}
\label{sec:method_gcgf}

\begin{figure}[tp]
    \centering
    \includegraphics[width=\linewidth]{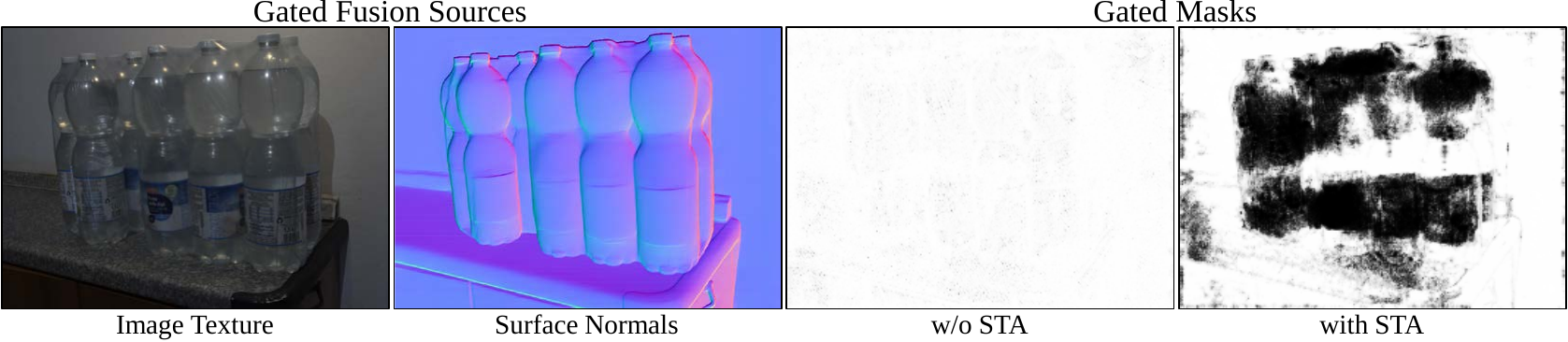}
    \caption{Comparison of gated mask effectiveness with and without Specular-Transparent Augmentation (STA). The STA significantly enhances the robustness of the gated mask in identifying the ambiguous image textures.}
    \label{fig:specular_transparent_augmentation_affect}
\end{figure}

To effectively fuse image and surface normal features for robust Syn-to-Real generalization, the \textbf{Gated Contextual-Geometric Fusion} comprises two primary components. The Gated Mask Network (GMNet) suppresses texture ambiguities as shown in Fig. \ref{fig:specular_transparent_augmentation_affect}, and the Fusion Network (FusionNet) merges the filtered image features with surface normal features, as depicted in Fig. \ref{fig:main_architecture}.

\textbf{Gated Mask Network.} Given the multi-scale image features $ \mathbf{f}_{I_{l(r)}}^{i} $ and surface normal features $ \mathbf{f}_{N_{l(r)}}^{i} $ at scales $ i \in \{4, 8, 16, 32\} $, together with the original stereo inputs $ \mathbf{I}_{l(r)} $ and $ \mathbf{N}_{l(r)} $, a shared-weight GMNet concatenates these modalities at the quarter-scale and then passes them through a standard convolution-instance-normalization-ReLU (ConvINReLU) pipeline to independently produce quarter-scale gated masks $ \mathbf{M}_{l(r)}^{4} $ for the left and right views, formulated as:

\begin{equation}
    \begin{aligned}
        \mathbf{f}_{l(r)}^{cat, 4} &= \mathrm{Concat}\{\mathbf{f}_{I_{l(r)}}^{4}, \mathbf{I}_{l(r)}^{4}, \mathbf{f}_{N_{l(r)}}^{4}, \mathbf{N}_{l(r)}^{4}\} \\
        \mathbf{f}_{\mathbf{M}_{l(r)}}^{4} &= \mathrm{ConvINReLU}^{4}(\mathbf{f}_{l(r)}^{cat, 4}) \\
        \mathbf{M}_{l(r)}^{4} &= \sigma(\mathrm{Conv}(\mathbf{f}_{\mathbf{M}_{l(r)}}^{4}))
    \end{aligned}
    \label{func:gmnet}
\end{equation}

\noindent where $ \sigma(\cdot) $ denotes the sigmoid function that constrains the gated mask values to the range $ [0, 1] $. Subsequently, $ \mathbf{M}_{l(r)}^{4} $ is downsampled to generate multi-scale gated masks $ \mathbf{M}_{l(r)}^{i} $ for $ i \in \{4, 8, 16, 32\} $. This ensures consistent suppression of image-texture ambiguities across all scales.

\textbf{Fusion Network.} Leveraging the multi-scale $ \mathbf{M}_{l(r)}^{i} $ for $ i \in \{4, 8, 16, 32\} $ from GMNet and $ \mathbf{f}_{I_{l(r)}}^{i} $, FusionNet performs an element-wise multiplication between them to obtain the multi-scale filtered image features $ \mathbf{f}_{I_{l(r)}}^{filter, i} $. Subsequently, $ \mathbf{f}_{I_{l(r)}}^{filter, i} $ is concatenated with $ \mathbf{f}_{N_{l(r)}}^{i} $ and processed through a ConvINReLU block at each scale to yield the multi-scale contextual-geometric features $ \mathbf{f}_{l(r)}^{fuse, i} $ as follows:

\begin{equation}
    \begin{aligned}
        \mathbf{f}_{I_{l(r)}}^{filter, i} &= \mathbf{f}_{I_{l(r)}}^{i} \odot \mathbf{M}_{l(r)}^{i} \\
        \mathbf{f}_{l(r)}^{cat, i} &= \mathrm{Concat}\{\mathbf{f}_{I_{l(r)}}^{filter, i}, \mathbf{f}_{N_{l(r)}}^{i}\} \\
        \mathbf{f}_{l(r)}^{fuse, i} &= \mathrm{ConvINReLU}^{i}(\mathbf{f}_{l(r)}^{cat, i})
    \end{aligned}
    \label{func:fusionnet}
\end{equation}

\noindent where $ \odot $ denotes the Hadamard Product. The $ \mathbf{f}_{l(r)}^{fuse, i} $ effectively suppress ambiguous texture features while integrating the complementary strengths of rich contextual image details and domain-invariant geometric cues from surface normals.

\subsection{Specular-Transparent Augmentation}
\label{sec:method_sta}

\begin{figure}[tp]
    \centering
    \includegraphics[width=\linewidth]{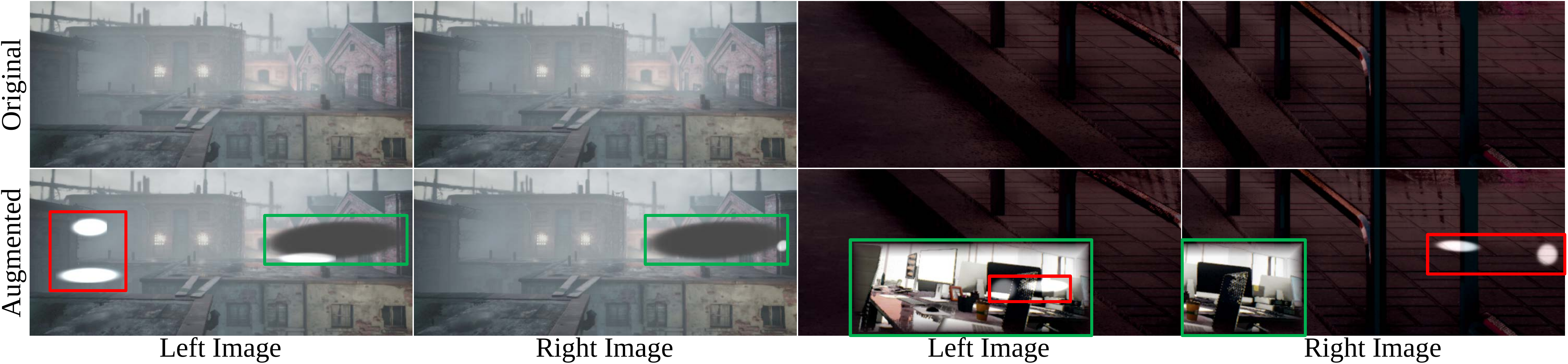}
    \caption{Visualization of Specular-Transparent Augmentation (STA). \textbf{Row 1}: Images without STA. \textbf{Row 2}: Images with STA, showcasing examples of specular (red boxes) and transparent (green boxes) augmentations.}
    \label{fig:specular_transparent_augmentation_visualization}
\end{figure}

Given that most synthetic datasets are generated under simplified illumination and trivial noise patterns (Fig. \ref{fig:comparison_of_images_and_normals} and \ref{fig:specular_transparent_augmentation_visualization}), the intrinsic left-right texture consistency in synthetic stereo pairs is insufficient for learning robust gated masks, especially for the non-Lambertian ambiguities in Fig. \ref{fig:specular_transparent_augmentation_affect}. To improve the robustness of the GCGF, the \textbf{Specular-Transparent Augmentation} (STA) strategy is proposed to intentionally perturb the consistency of image textures and generate more complex texture appearances during training. As shown in Fig. \ref{fig:specular_transparent_augmentation_visualization}, STA comprises two core components, which are Specular Augmentation (SA) and Transparent Augmentation (TA).

\textbf{Specular Augmentation.} To simulate texture inconsistencies caused by view-dependent highlight shifts under specular reflection (e.g., the "Bottle Surface" in the Image Texture of Fig. \ref{fig:specular_transparent_augmentation_affect}), Specular Augmentation (SA) randomly selects one or more ellipsoidal regions in the left and right views and assigns white RGB values to these areas to emulate highlight characteristics. To improve the realism of the synthesized highlights, SA further randomly adjusts the intensity of the white RGB values and applies Gaussian blur to the ellipsoidal boundaries, thereby approximating the scattering behavior observed along real physical highlight edges, as illustrated in Fig. \ref{fig:specular_transparent_augmentation_visualization}.

\textbf{Transparent Augmentation.} In realistic scenarios, transparent and mirror-like objects, such as glass and mirrors, exhibit \textit{imaginary textures} caused by the superposition of intrinsic textures and reflected background or foreground content. These textures are typically consistent between the left and right views. To simulate the ambiguities introduced by this phenomenon, Transparent Augmentation (TA) first randomly selects a region in the left image and then determines its corresponding region in the right image using the median ground-truth disparity of the selected left region. Subsequently, TA replaces the textures within the selected regions in both left and right images with content sampled from another image in the dataset. To account for slight inconsistencies induced by viewpoint differences, the substituted texture in the right image is randomly shifted along the epipolar line. In addition, to emulate textureless areas, TA randomly fills the selected regions with gray RGB values, as shown in Fig. \ref{fig:specular_transparent_augmentation_visualization}.

\subsection{Sparse Global Aggregation}
\label{sec:method_sparse}

To preserve the robust capability of GREAT-Stereo \cite{greatstereo} in addressing ill-posedness within occluded, textureless, and repetitive regions, GREATEN-Stereo retains its global feature extraction modules, while substantially reducing the computational overhead of the designs in \cite{greatstereo}. Specifically, GREATEN extracts multi-scale global contextual-geometric features from $ \mathbf{f}_{l(r)}^{fuse, i} $ for $ i \in \{4, 8, 16, 32\} $ via Sparse Deformable Attentions \cite{deformdetr, deformvit, bevformer, gaussianformer, gaussianformerv2}, as demonstrated in Fig. \ref{fig:main_architecture}.

\textbf{Sparse Spatial Attention.} To enhance object structure propagation within occluded regions, the \textbf{Sparse Spatial Attention} (SSA) utilizes 2D deformable attention \cite{gaussianformer, gaussianformerv2} to extract global contextual-geometric details from multi-scale left features $ \mathbf{f}_{l}^{fuse, i} \in \mathbb{R}^{C_i \times \frac{H}{i} \times \frac{W}{i}} $, as depicted in Fig. \ref{fig:main_architecture}. Specifically, the \textbf{Key Points Sampling} (KPS) module processes $ \mathbf{f}_{l}^{fuse, i} $ through a linear layer to predict $ k $ sampling points $ \mathbf{p}_{l}^{i} \in \mathbb{R}^{2 \times k \times \frac{H}{i} \times \frac{W}{i}} $ across the entire spatial domain for each pixel, where the dimension "2" corresponds to horizontal (x) and vertical (y) coordinates. Subsequently, KPS employs $ \mathbf{p}_{l}^{i} $ to sample corresponding feature vectors $ \mathbf{v}_{l}^{fuse, i} \in \mathbb{R}^{k \times C_i \times \frac{H}{i} \times \frac{W}{i}} $ from $ \mathbf{f}_{l}^{fuse, i} $ for each pixel, formulated as:

\begin{equation}
    \begin{aligned}
        \mathbf{p}_{l}^{i} &= \mathrm{Linear}^{i}(\mathbf{f}_{l}^{fuse, i}) \\
        \mathbf{v}_{l}^{fuse, i} &= \mathcal{G}(\mathrm{Linear}^{i}(\mathbf{f}_{l}^{fuse, i}), \mathbf{p}_{l}^{i})
    \end{aligned}
    \label{func:kps}
\end{equation}

\noindent where $ \mathcal{G}(A, B) $ denotes the grid sampling function that extracts values from $ A $ utilizing key points $ B $, and $ \mathrm{Linear} $ represents a linear layer. Concurrently, the \textbf{Global Aggregation} (GA) module passes $ \mathbf{f}_{l}^{fuse, i} $ through another linear layer to generate attention weights $ W_{l}^{attn, i} \in \mathbb{R}^{k \times 1 \times \frac{H}{i} \times \frac{W}{i}} $ corresponding to each sampled vector. Finally, global feature aggregation is achieved via a weighted sum as follows:

\begin{equation}
    \begin{aligned}
        \mathbf{f}_{l}^{fuse, i} &= \mathrm{PE}(\mathbf{f}_{l}^{fuse, i}) + \mathbf{f}_{l}^{fuse, i} \\
        \mathbf{v}_{l}^{fuse, i} &= \mathrm{KPS}(\mathbf{f}_{l}^{fuse, i}) \\
        \mathbf{W}_{l}^{attn, i} &= \mathrm{Softmax}(\mathrm{Linear}^{i}(\mathbf{f}_{l}^{fuse, i})) \\
        \mathbf{f}_{l}^{spatial, i} &= \mathrm{LN}(\mathbf{f}_{l}^{fuse, i} + \mathrm{Linear}^{i}(\mathbf{W}_{l}^{attn, i} \cdot \mathbf{v}_{l}^{fuse, i})) \\
        \mathbf{f}_{l}^{spatial, i} &= \mathrm{LN}(\mathbf{f}_{l}^{spatial, i} + \mathrm{Linear}^{i}(\mathbf{f}_{l}^{spatial, i}))
    \end{aligned}
    \label{func:ga}
\end{equation}

\noindent where PE denotes positional embedding \cite{transformer, vit} and LN stands for LayerNorm \cite{layernorm}.

\textbf{Sparse Dual-Matching Attention.} To mitigate matching ambiguities in textureless and repetitive regions while encoding domain-invariant geometric cues from surface normals, \textbf{Sparse Dual-Matching Attention} (SDMA) captures global contextual and geometric details along epipolar lines. This is achieved via \textbf{Bi-Directional Epipolar Sparse Cross-Attention} (BESC), which leverages quarter-scale filtered image features and surface normal features from both views ($ \mathbf{f}_{I_{l(r)}}^{filter, 4} $ and $ \mathbf{f}_{N_{l(r)}}^{4} \in \mathbb{R}^{C_4 \times \frac{H}{4} \times \frac{W}{4}} $), as illustrated in Fig. \ref{fig:main_architecture}. By using contextual-geometric features $ \mathbf{f}_{l(r)}^{fuse, 4} $ from the GCGF module (Sec. \ref{sec:method_gcgf}) as queries, alongside $ \mathbf{f}_{I_{l(r)}}^{filter, 4} $ and $ \mathbf{f}_{N_{l(r)}}^{4} $ as keys and values, BESC generates matching left and right features ($ \mathbf{f}_{l(r)}^{mat, 4} \in \mathbb{R}^{C_4 \times \frac{H}{4} \times \frac{W}{4}} $) for robust cost volume construction. To expedite bi-directional cross-attention between the left and right views, these features are concatenated along the batch axis, formulated as follows:

\begin{equation}
    \begin{aligned}
        \mathbf{f}_{l \rightarrow r}^{fuse, 4} &= \mathrm{Concat}\{\mathbf{f}_{l}^{fuse, 4}, \mathbf{f}_{r}^{fuse, 4}\} \\
        \mathbf{f}_{I_{r \rightarrow l}}^{filter, 4} &= \mathrm{Concat}\{\mathbf{f}_{I_r}^{filter, 4}, \mathbf{f}_{I_l}^{filter, 4}\} \\
        \mathbf{f}_{N_{r \rightarrow l}}^{4} &= \mathrm{Concat}\{\mathbf{f}_{N_r}^{4}, \mathbf{f}_{N_l}^{4}\}
    \end{aligned}
    \label{func:concate}
\end{equation}

\noindent Subsequently, BESC utilizes $ \mathbf{f}_{l \rightarrow r}^{fuse, 4} $ to generate $ k $ sampling points $ \mathbf{p}_{r \rightarrow l}^{4} \in \mathbb{R}^{2 \times k \times \frac{H}{4} \times \frac{W}{4}} $ for $ \mathbf{f}_{I_{r \rightarrow l}}^{filter, 4} $ and $ \mathbf{f}_{N_{r \rightarrow l}}^{4} $, respectively, by adopting the KPS module from SSA. To constrain the sampling points along epipolar lines, the vertical (y) coordinates of each point are set identical to the pixel coordinates. Analogous to SSA, $ \mathbf{p}_{r \rightarrow l}^{4} $ are then employed to extract corresponding feature vectors ($ \mathbf{v}_{I_{r \rightarrow l}}^{filter, 4} $ and $ \mathbf{v}_{N_{r \rightarrow l}}^{4} $) from $ \mathbf{f}_{I_{r \rightarrow l}}^{filter, 4} $ and $ \mathbf{f}_{N_{r \rightarrow l}}^{4} $ as follows:

\begin{equation}
    \begin{aligned}
        \mathbf{f}_{l \rightarrow r}^{fuse, 4} &= \mathrm{PE}(\mathbf{f}_{l \rightarrow r}^{fuse, 4}) + \mathbf{f}_{l \rightarrow r}^{fuse, 4} \\
        \mathbf{v}_{I_{r \rightarrow l}}^{filter, 4} &= \mathrm{KPS}_{I_{r \rightarrow l}}(\mathrm{LN}(\mathbf{f}_{l \rightarrow r}^{fuse, 4}), \mathrm{LN}(\mathbf{f}_{I_{r \rightarrow l}}^{filter, 4})) \\
        \mathbf{v}_{N_{r \rightarrow l}}^{4} &= \mathrm{KPS}_{N_{r \rightarrow l}}(\mathrm{LN}(\mathbf{f}_{l \rightarrow r}^{fuse, 4}), \mathrm{LN}(\mathbf{f}_{N_{r \rightarrow l}}^{4}))
    \end{aligned}
    \label{func:besc_kps}
\end{equation}

\noindent where $ \mathrm{KPS}_{*}(A, B) $ denotes the key points sampling operation that generates key points from $ A $ to sample features from $ B $, with $ * $ referring to $ I_{r \rightarrow l} $ or $ N_{r \rightarrow l} $. Finally, global feature aggregation is achieved via a weighted sum as follows:

\begin{equation}
    \begin{aligned}
        \mathbf{W}_{I_{r \rightarrow l}}^{attn} &= \mathrm{Softmax}(\mathrm{Linear}_{I_{r \rightarrow l}}(\mathrm{LN}(\mathbf{f}_{l \rightarrow r}^{fuse, 4}))) \\
        \mathbf{W}_{N_{r \rightarrow l}}^{attn} &= \mathrm{Softmax}(\mathrm{Linear}_{N_{r \rightarrow l}}(\mathrm{LN}(\mathbf{f}_{l \rightarrow r}^{fuse, 4}))) \\
        \mathbf{f}_{temp} &= \mathrm{Concat}\{\mathbf{W}_{I_{r \rightarrow l}}^{attn} \cdot \mathbf{v}_{I_{r \rightarrow l}}^{filter, 4}, \mathbf{W}_{N_{r \rightarrow l}}^{attn} \cdot \mathbf{v}_{N_{r \rightarrow l}}^{4}\} \\
        \mathbf{f}_{l \rightarrow r}^{mat, 4} &= \mathbf{f}_{l \rightarrow r}^{fuse, 4} + \mathrm{Linear}(\mathbf{f}_{temp}) \\
        \mathbf{f}_{l \rightarrow r}^{mat, 4} &= \mathbf{f}_{l \rightarrow r}^{mat, 4} + \mathrm{Linear}(\mathrm{LN}(\mathbf{f}_{l \rightarrow r}^{mat, 4}))
    \end{aligned}
    \label{func:besc}
\end{equation}

\noindent where $ \mathbf{W}_{*}^{attn} \in \mathbb{R}^{k \times 1 \times \frac{H}{4} \times \frac{W}{4}} $ represents the attention weights, with $ * $ referring to $ I_{r \rightarrow l} $ or $ N_{r \rightarrow l} $. Subsequently, $ \mathbf{f}_{l \rightarrow r}^{mat, 4} $ is divided into $ \mathbf{f}_{l}^{mat, 4} $ and $ \mathbf{f}_{r}^{mat, 4} $ along the batch axis for subsequent processing.

\textbf{Simple Volume Attention.} Observing that the core function of Volume Attention (VA) in GREAT-Stereo \cite{greatstereo} is to regularize the 4D Cost Volume ($ \mathbf{C}_{cv} \in \mathbb{R}^{C_{4} \times D \times \frac{H}{4} \times \frac{W}{4}} $) across both the spatial dimensions ($ \frac{H}{4} \times \frac{W}{4} $) and the disparity search space ($ C_{4} \times D $), where $ d \in \{0, 1, \dots, D - 1\} $ denotes the $ d^{th} $ disparity candidate. The \textbf{Simple Volume Attention} (SVA) is proposed to replace the complex attention mechanism in VA with a 4D volume filter ($ \mathbf{C}_{filter} \in \mathbb{R}^{N_{g} \times D \times \frac{H}{4} \times \frac{W}{4}} $). This filter is generated by combining the cost volume with the features derived from SSA ($ \mathbf{f}_{l}^{spatial, 4} $) and then used to regularize the cost volume. Specifically, the combined cost volume ($ \mathbf{C}_{comb} \in \mathbb{R}^{N_{g} \times D \times \frac{H}{4} \times \frac{W}{4}} $) is constructed as follows:

\begin{equation}
    \begin{aligned}
        \mathbf{C}_{cor}(g, d, x, y) &= \alpha \cdot \langle \mathbf{f}_{l}^{mat, 4}(x, y), \mathbf{f}_{r}^{mat, 4}(x - d, y) \rangle \\
        \mathbf{f}_{l(r)}^{mat, 4} &= \mathrm{Conv}(\mathbf{f}_{l(r)}^{mat, 4}) \\
        \mathbf{C}_{cat}(2g, d, x, y) &= \langle \mathbf{f}_{l}^{mat, 4}(x, y))\ |\ \mathbf{f}_{r}^{mat, 4}(x - d, y) \rangle \\
        \mathbf{C}_{comb}(g, d, x, y) &= \mathrm{Conv}(\langle \mathbf{C}_{corr} \ |\ \mathbf{C}_{cat} \rangle)
    \end{aligned}
    \label{func:combined_volume}
\end{equation}

\noindent where $ \alpha = \frac{1}{C_{4}/N_{g}} $ and $ N_{g} = 8 $ denotes the number of groups along the channel dimension to compute correlation maps group by group. $ \langle \cdot, \cdot \rangle $ and $ \langle \cdot |\cdot \rangle $ denote the inner product and concatenation along channel dimension, respectively. $ \mathbf{C}_{cor} $ and $ \mathbf{C}_{cat} $ represent the correlation volume and the concatenation volume, respectively. Subsequently, $ \mathrm{C}_{filter} $ and the refined cost volume $ \mathbf{C}_{comb}^{refine} $ are formulated as:

\begin{equation}
    \begin{aligned}
        \mathbf{C}_{filter} &= \mathrm{Softmax}(\mathrm{Conv}(\mathbf{C}_{comb})) \cdot \mathrm{Conv}(\mathbf{f}_{l}^{spatial, 4}) \\
        \mathbf{C}_{comb}^{refine} &= \sigma(\mathrm{C}_{filter}) \cdot \mathbf{C}_{comb}
    \end{aligned}
    \label{func:refined_combined_volume}
\end{equation}

\noindent where $ \sigma(\cdot) $ represents the sigmoid function. This formulation enables SVA to retain the capability of VA \cite{greatstereo} to regularize the cost volume, while substantially reducing the computational overhead inherent in full attention mechanisms.

\subsection{VFM-Enhanced GREATEN framework}
\label{sec:method_vfm_enhanced}

For a fair comparison with models built upon Vision Foundation Models (VFMs) \cite{monsterstereo, defomstereo, foundationstereo}, we further extend GREATEN into a VFM-Enhanced variant by incorporating the widely adopted DepthAnythingV2 \cite{depthanythingv2}. Following the designs of FoundationStereo \cite{foundationstereo} and DEFOM-Stereo \cite{defomstereo}, the VFM-Enhanced GREATEN framework leverages both image features and relative monocular depth priors derived from VFMs.

\textbf{Image Feature Priors.} Pretrained on large-scale datasets, DepthAnythingV2 provides image feature priors with rich textural context, which can facilitate the Fusion Network of GCGF (\ref{sec:method_gcgf}) and the Context Network of iterative methods \cite{raftstereo, igevstereo} in capturing object-level contextual information. To preserve these effective priors, both GCGF and the Context Network take these image feature priors as auxiliary inputs and concatenate them with the original image features, yielding a VFM-Enhanced image feature representation.

\textbf{Relative Monocular Depth Priors.} Beyond image feature priors, DepthAnythingV2 \cite{depthanythingv2} also provides a relative monocular depth prior with strong geometric structure and consistency. To exploit this prior for accelerating the ConvGRU iterations, the \textbf{Learnable Scale Shift} module takes the initial disparity ($ \mathbf{d}_{0} $) regressed from $ \mathrm{C}_{comb}^{refine} $ and the relative depth prior ($ \mathbf{d}_{rel} $) as inputs, aligning $ \mathbf{d}_{rel} $ to the metric scale as follows:

\begin{equation}
    \begin{aligned}
        \alpha, \beta &= \mathrm{ConvBlock}(\mathrm{Concat}\{\mathbf{d}_{0}, \mathbf{d}_{rel}\}) \\
        \mathbf{d}_{met} &= \alpha \cdot \mathbf{d}_{rel} + \beta
    \end{aligned}
    \label{func:learnable_scale_shift}
\end{equation}

\noindent where $ \alpha $ and $ \beta $ denote the alignment scalars, and $ \mathbf{d}_{met} $ represents the metric disparity used to initialize the ConvGRU.

\subsection{Loss Function}
\label{sec:method_loss}

Following IGEV-Stereo \cite{igevstereo}, the initial disparity $ \mathbf{d}_{0} $ regressed from $ \mathbf{C}_{comb}^{refine} $ and the metric disparity $ \mathbf{d}_{met} $ are supervised by the Smooth L1 loss \cite{psmnet}, if applicable. Meanwhile, all subsequent updated disparities $ \{\mathbf{d}_{i=1}^{N}\} $ are optimized using an L1 loss with exponentially increasing weights \cite{raftstereo}. The total loss is defined as follows:

\begin{equation}
    \begin{aligned}
        \mathcal{L}_{init} &= \mathrm{Smooth}_{\mathrm{L_1}}(\mathbf{d}_{0} - \mathbf{d}_{gt}) \\
        \mathcal{L}_{met} &= \mathrm{Smooth}_{\mathrm{L_1}}(\mathbf{d}_{met} - \mathbf{d}_{gt}) \\
        \mathcal{L}_{stereo} &= \mathcal{L}_{init} + \mathcal{L}_{met} +\Sigma_{i=1}^{N}\gamma^{N-i}||\mathbf{d}_{i} - \mathbf{d}_{gt}||_{1}
    \end{aligned}
    \label{func:loss}
\end{equation}
\noindent where $ \mathbf{d}_{gt} $ denotes the ground truth disparity and $ \gamma = 0.9 $.

\begin{figure*}[tp]
    \centering
    \includegraphics[width=0.97\linewidth]{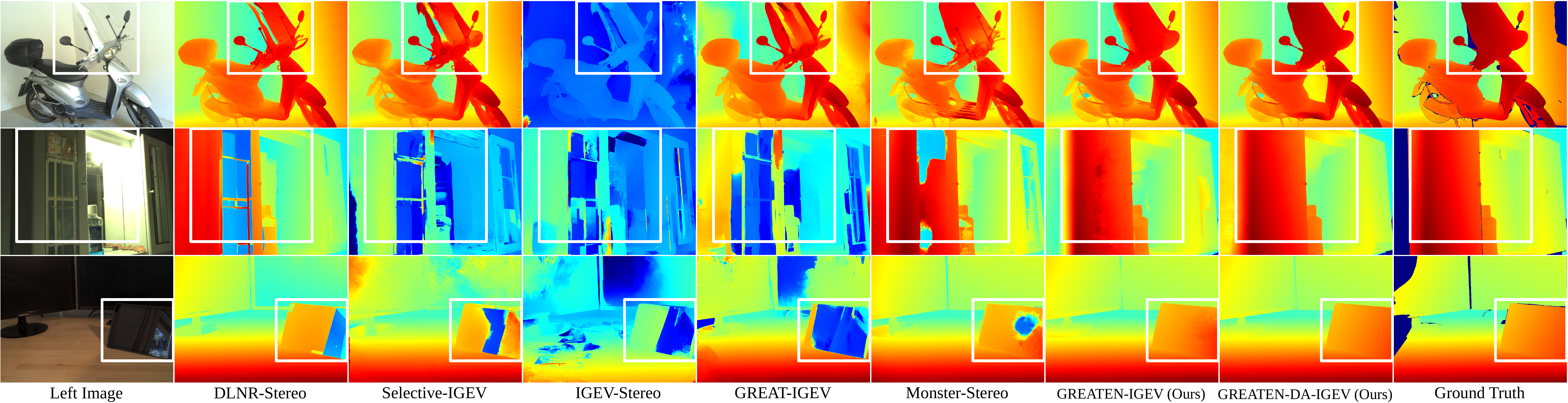}
    \caption{Zero-Shot qualitative results on non-Lambertian Booster \cite{booster} training set. Our GREATEN-IGEV and GREATEN-DepthAny-IGEV outperform other iterative methods, where "DA" stands for "DepthAny". All the models are trained exclusively on Synthetic datasets.}
    \label{fig:booster_zero_shot_qualitative_results}
\end{figure*}

\begin{figure*}[tp]
    \centering
    \includegraphics[width=0.97\linewidth]{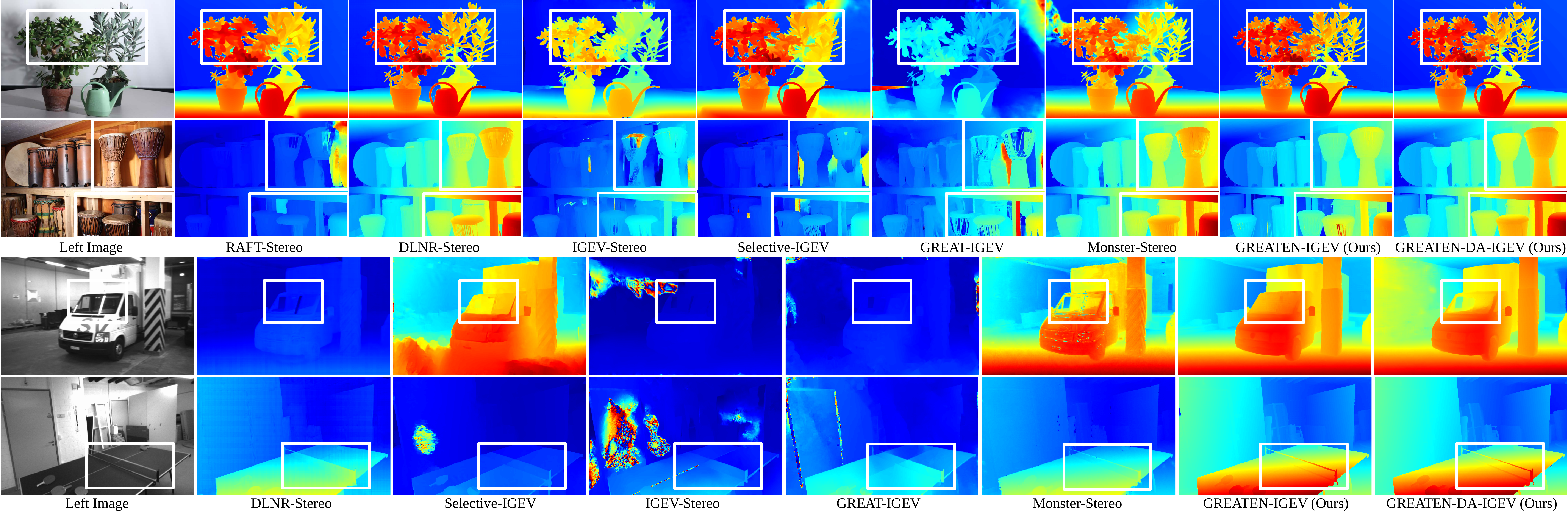}
    \caption{Zero-Shot qualitative results on Middlebury \cite{middlebury} (\textbf{Row 1 \& 2}) and ETH3D \cite{eth3d} (\textbf{Row 3 \& 4}) testing set. Our GREATEN-IGEV and GREATEN-DepthAny-IGEV outperform other iterative methods, where "DA" stands for "DepthAny". All the models are trained exclusively on Synthetic datasets.}
    \label{fig:eth3d_middlebury_zero_shot_qualitative_results}
\end{figure*}

\section{Experiment}
\label{sec:experiment}

\subsection{Datasets and Evaluation Metrics}
\label{sec:datasets}

\textbf{Datasets.} We utilize one synthetic and five real-world datasets for the basic evaluation. \textbf{SceneFlow} \cite{sceneflow} is a synthetic dataset containing 35454 training and 4370 testing pairs annotated with dense disparity maps. We adopt its finalpass version, as it is more realistic and challenging than the cleanpass. \textbf{KITTI-2012} \cite{kitti2012} and \textbf{KITTI-2015} \cite{kitti2015} provide real-world driving scene data. KITTI-2012 contains 194 training and 195 testing pairs, while KITTI-2015 offers 200 training and 200 testing pairs. \textbf{ETH3D} \cite{eth3d} comprises 27 training and 20 testing pairs of gray-scale stereo images spanning both indoor and outdoor realistic scenarios. \textbf{Middlebury} \cite{middlebury} consists of high-resolution realistic indoor scenes featuring 15 training and 15 testing pairs. \textbf{Booster} \cite{booster} provides 228 training pairs with high-quality disparity labels for high-resolution indoor scenes. Notably, Booster features specular and transparent surfaces, which are the primary causes of synthetic-to-realistic (Syn-to-Real) generalization failures in modern stereo matching.

\textbf{Evaluation Metrics.} We evaluate the GREATEN framework using two standard metrics. End-Point Error (EPE) is defined as the mean absolute disparity error, and the $ \mathrm{x} $-pixel error rate (D$ \mathrm{x} $) denotes the proportion of pixels whose disparity error exceeds $ \mathrm{x} $ pixels. When applicable, we report these metrics over all dense pixels (\textit{All}), non-occluded dense pixels (\textit{Noc}), and occluded dense pixels (\textit{Occ}).

\begin{table*}[!t]
    \caption{Syn-to-Real Zero-Shot generalization on KITTI and Booster datasets. Unless specified, all models are trained from scratch on SceneFlow \cite{sceneflow}. StereoAnywhere \textsuperscript{\ddag} is trained from a frozen RAFT-Stereo checkpoint with extra supervision for surface normals, and uses priors of VFM pretrained on the HyperSim \cite{hypersim} dataset. \textsuperscript{*} denotes training on our Syn-to-Real Mixed datasets.}
    \label{tab:zero_shot_for_kitti_and_booster}
    \centering
    \resizebox{1.98\columnwidth}{!}{
        \begin{tabular}{l|cccc|cccc|ccccc}
            \toprule
            \multirow{3}{*}{Model} & \multicolumn{4}{c|}{KITTI-2012 \cite{kitti2012}} & \multicolumn{4}{c|}{KITTI-2015 \cite{kitti2015}} & \multicolumn{5}{c}{Booster (Q) \cite{booster}} \\
            \cmidrule{2-14}
            & \multicolumn{2}{c}{EPE $ \downarrow $} & \multicolumn{2}{c|}{D3 $ \downarrow $}& \multicolumn{2}{c}{EPE $ \downarrow $} & \multicolumn{2}{c|}{D3 $ \downarrow $} & \multirow{2}{*}{EPE $ \downarrow $} & \multirow{2}{*}{D2 $ \downarrow $} & \multirow{2}{*}{D4 $ \downarrow $} & \multirow{2}{*}{D6 $ \downarrow $} & \multirow{2}{*}{D8 $ \downarrow $} \\
            & All & Noc & All & Noc & All & Noc & All & Noc & & & & \\
            \hline
            \multicolumn{14}{c}{VFM-Free} \\
            \hline
            RAFT-Stereo \cite{raftstereo} & \trd 0.92 & \trd 0.74 & \trd 4.65 & 2.84 & \snd 1.12 & 0.97 & 5.75 & 4.45 & \trd 3.59 & 17.84 & 13.06 & 10.76 & 9.24 \\
            PCV-Stereo \cite{pcvstereo} & 0.95 & 0.78 & 4.73 & 3.12 & 1.34 & 1.19 & \trd 5.73 & 4.67 & 4.70 & 22.63 & 16.51 & 13.81 & 12.08 \\
            DLNR-Stereo \cite{dlnrstereo} & 1.59 & 1.44 & 9.09 & 6.24 & 2.83 & 2.82 & 16.07 & 13.71 & 3.97 & 18.56 & 14.55 & 12.61 & 11.22 \\
            IGEV-Stereo \cite{igevstereo} & 1.03 & 0.82 & 5.13 & 3.08 & 1.21 & 1.02 & 6.03 & 4.56 & 4.19 & \snd 17.47 & 13.79 & 12.05 & 10.88 \\
            Selective-IGEV \cite{selectivestereo} & 1.08 & 0.92 & 5.64 & 3.47 & 1.25 & 1.08 & 6.05 & 4.75 & 4.38 & 18.52 & 14.24 & 12.14 & 10.77 \\
            GREAT-IGEV \cite{greatstereo} & 1.02 & 0.81 & 5.31 & 3.35 & 1.16 & 0.99 & 5.88 & 4.41 & 4.61 & 19.30 & 14.60 & 13.51 & 12.10 \\
            GREATEN-Selective (Ours) & \trd 0.92 & \snd 0.72 & 4.67 & \trd 2.39 & \trd 1.15 & \trd 0.96 & 5.79 & \trd 4.12 & \snd 2.85 & 17.76 & \trd 11.34 & \trd 8.62 & \snd 7.00 \\
            GREATEN-IGEV (Ours) & \snd 0.91 & \snd 0.72 & \snd 4.54 & \snd 2.27 & \snd 1.12 & \snd 0.94 & \snd 5.70 & \snd 4.09 & \snd 2.85 & \trd 17.55 & \snd 11.10 & \snd 8.47 & \trd 7.09 \\
            GREATEN-IGEV\textsuperscript{*} (Ours) & \fst 0.73 & \fst 0.56 & \fst 3.30 & \fst 1.50 & \fst 0.93 & \fst 0.80 & \fst 3.75 & \fst 2.38 & \fst \textbf{1.97} & \fst \textbf{11.95} & \fst \textbf{7.45} & \fst \textbf{5.67} & \fst \textbf{4.46} \\
            GREATEN-Selective\textsuperscript{*} (Ours) & \fst \textbf{0.71} & \fst \textbf{0.54} & \fst \textbf{3.21} & \fst \textbf{1.46} & \fst \textbf{0.89} & \fst \textbf{0.77} & \fst \textbf{3.47} & \fst \textbf{2.16} & \fst 2.40 & \fst 12.92 & \fst 8.36 & \fst 6.62 & \fst 5.57 \\
            \hline
            \multicolumn{14}{c}{VFM-Enhanced} \\
            \hline
            StereoAnywhere\textsuperscript{\ddag} \cite{stereoanywhere} & 0.83 & - & 3.90 & 3.52 & 0.97 & - & 3.93 & 3.79 & 1.21 & 9.01 & 5.40 & 4.12 & 3.34 \\
            \hline
            DEFOM-Stereo \cite{defomstereo} & \snd 0.83 & 0.78 & \snd 3.90 & 3.52 & \fst 1.06 & \snd 1.04 & 4.99 & 4.76 & 2.92 & \trd 12.77 & 9.53 & 8.22 & 7.42 \\
            MG-Stereo \cite{mgstereo} & 0.86 & 0.80 & 4.14 & 3.71 & \trd 1.12 & 1.10 & 5.64 & 5.40 & \fst 2.14 & \fst 9.97 & \fst 6.84 & \fst 5.75 & \fst 5.23 \\
            Bridge-Depth \cite{bridgedepth} & \snd 0.83 & \trd 0.77 & \trd 4.04 & 3.63 & \snd 1.09 & 1.07 & \snd 4.69 & \trd 4.52 & 3.13 & 14.74 & 9.71 & 8.01 & 7.03 \\
            Monster-Stereo \cite{monsterstereo} & 0.93 & 0.88 & 4.62 & 4.12 & 1.17 & 1.15 & 5.52 & 5.32 & 3.04 & 19.17 & 11.49 & 8.25 & 7.69 \\
            SMoE-Stereo \cite{smoestereo} & \snd 0.83 & \trd 0.77 & 4.23 & 3.76 & \fst 1.06 & \snd 1.04 & \trd 4.94 & 4.77 & 2.80 & 20.81 & 11.80 & 8.65 & 6.85 \\
            Prompt-Stereo \cite{promptstereo} & \fst 0.79 & \snd 0.73 & \fst 3.77 & \trd 3.32 & \snd 1.09 & \trd 1.06 & \fst 4.59 & \snd 4.39 & \trd 2.78 & \snd 12.13 & \trd 8.93 & \trd 7.54 & \trd 6.79 \\
            GREATEN-DepthAny-IGEV (Ours) & \trd 0.84 & \fst 0.66 & \snd 3.90 & \fst 2.00 & \fst 1.06 & \fst 0.89 & 5.06 & \fst 3.51 & \snd 2.29 & 13.99 & \snd 8.30 & \snd 6.32 & \snd 5.29 \\
            GREATEN-DepthAny-IGEV\textsuperscript{*} (Ours) & \fst \textbf{0.72} & \fst \textbf{0.55} & \fst \textbf{3.11} & \fst \textbf{1.44} & \fst \textbf{0.91} & \fst \textbf{0.78} & \fst \textbf{3.61} & \fst \textbf{2.33} & \fst \textbf{1.53} & \fst \textbf{9.81} & \fst \textbf{5.89} & \fst \textbf{4.67} & \fst \textbf{3.93} \\
            \bottomrule
        \end{tabular}
    }
\end{table*}

\begin{table}[!t]
    \caption{Syn-to-Real Zero-Shot generalization on Non-occluded ETH3D and Middlebury. Unless specified, all models are trained on SceneFlow \cite{sceneflow}. "DA" represents "DepthAny". \textsuperscript{*} denotes training on our Syn-to-Real Mixed datasets.}
    \label{tab:zero_shot_for_eth3d_and_middlebury}
    \centering
    \resizebox{0.98\columnwidth}{!}{
        \begin{tabular}{l|cc|cc}
            \toprule
            \multirow{2}{*}{Model} & \multicolumn{2}{c|}{ETH3D \cite{eth3d}} & \multicolumn{2}{c}{Middlebury (H) \cite{middlebury}} \\
            & EPE $ \downarrow $ & D1 $ \downarrow $ & EPE $ \downarrow $ & D2 $ \downarrow $ \\
            \hline
            \multicolumn{5}{c}{VFM-Free} \\
            \hline
            RAFT-Stereo \cite{raftstereo} & 0.31 & 3.22 & 2.25 & 10.04 \\
            IGEV-Stereo \cite{igevstereo} & 0.29 & 3.61 & 2.28 & 9.48 \\
            Selective-IGEV \cite{selectivestereo} & 1.36 & 5.40 & 2.31 & 9.22 \\
            GREAT-IGEV \cite{greatstereo} & \trd 0.28 & 3.80 & 2.40 & 8.58 \\
            GREATEN-Selective (Ours) & \snd 0.22 & \trd 2.48 & \trd 1.84 & \trd 7.88 \\
            GREATEN-IGEV (Ours) & \fst 0.18 & \snd 1.26 & \snd 0.77 & \snd 6.63 \\
            GREATEN-IGEV\textsuperscript{*} (Ours) & \fst \textbf{0.16} & \fst \textbf{1.05} & \fst \textbf{0.70} & \fst 5.23 \\
            GREATEN-Selective\textsuperscript{*} (Ours) & \fst \textbf{0.16} & \fst 1.08 & \fst 0.75 & \fst \textbf{4.81} \\
            \hline
            \multicolumn{5}{c}{VFM-Enhanced} \\
            \hline
            FoundationStereo \cite{foundationstereo} & - & \snd 1.80 & - & \ \snd 5.50 \\
            Monster-Stereo \cite{monsterstereo} & 0.26 & 2.86 & 0.94 & 7.27 \\
            Monster++ \cite{monster++} & - & \trd 2.03 & - & \fst 5.23 \\
            SMoE-Stereo \cite{smoestereo} & 0.22 & 2.08 & 1.29 & 7.06 \\
            GREATEN-DA-IGEV (Ours) & \fst 0.19 & \fst 1.37 & \fst 0.81 & \trd 6.51 \\
            GREATEN-DA-IGEV\textsuperscript{*} (Ours) & \fst \textbf{0.16} & \fst \textbf{0.93} & \fst \textbf{0.55} & \fst \textbf{3.54} \\
            \bottomrule
        \end{tabular}
    }
\end{table}

\subsection{Implementation Details}
\label{sec:implementation_details}

\textbf{Model Variants.} To demonstrate universality, we instantiate our framework on IGEV-Stereo \cite{igevstereo} and Selective-IGEV \cite{selectivestereo} to yield two \textbf{VFM-Free} variants: \textbf{GREATEN-IGEV} and \textbf{GREATEN-Selective}. In these variants, surface normals are treated merely as a generic input modality, obtainable from any off-the-shelf estimator. Meanwhile, the model architecture is completely decoupled from and agnostic to the Vision Foundation Models (VFMs). For fair comparison with VFM-based methods, we introduce a \textbf{VFM-Enhanced} variant, \textbf{GREATEN-DepthAny-IGEV}, which explicitly leverages frozen features and depth priors from DepthAnythingV2 \cite{depthanythingv2} for auxiliary features and ConvGRU initialization (Sec. \ref{sec:method_vfm_enhanced}). We collectively term these \textbf{GREATEN-Stereo}.


\textbf{Training Details.} We implement GREATEN-Stereo in PyTorch and train models on NVIDIA RTX 4090 GPUs. In all experiments, we employ the AdamW optimizer \cite{adamw} with gradient clipping set to [-1, 1] and a one-cycle learning rate scheduler \cite{onecycle} with an initial learning rate of 2e-4. Pretraining is performed from scratch on the SceneFlow \cite{sceneflow} dataset for 200k steps with a batch size of 8. To improve synthetic-to-realistic (\textbf{Syn-to-Real}) generalization, we further train the model on a mixed synthetic dataset comprising SceneFlow \cite{sceneflow}, FallingThings \cite{fallingthings}, TartanAir \cite{tartanair}, CREStereo \cite{crestereo}, Sintel \cite{sintel}, CARLA-HRVS \cite{hrvs}, and Virtual KITTI 2 \cite{vkitti2} for an additional 200k steps with a batch size of 8, initialized from the SceneFlow checkpoint. For \textbf{Robust Vision Challenge} (RVC) fine-tuning, we train the model on five standard real-world datasets (KITTI-2012 \cite{kitti2012}, KITTI-2015 \cite{kitti2015}, Middlebury \cite{middlebury}, ETH3D \cite{eth3d}, and InStereo2k \cite{instereo2k}) for 100k additional steps with the mixed Syn-to-Real checkpoint. We apply standard data augmentations following RAFT-Stereo \cite{raftstereo} with our Specular-Transparent Augmentation (STA) and a crop size of 384$ \times $768. We use 22 iterations during training and 32 iterations during inference. The number of attention blocks follows the standard settings in GREAT-Stereo \cite{greatstereo}. Surface normals are obtained following the pipeline in StereoAnywhere \cite{stereoanywhere}.

\begin{figure}[tp]
    \centering
    \includegraphics[width=\linewidth]{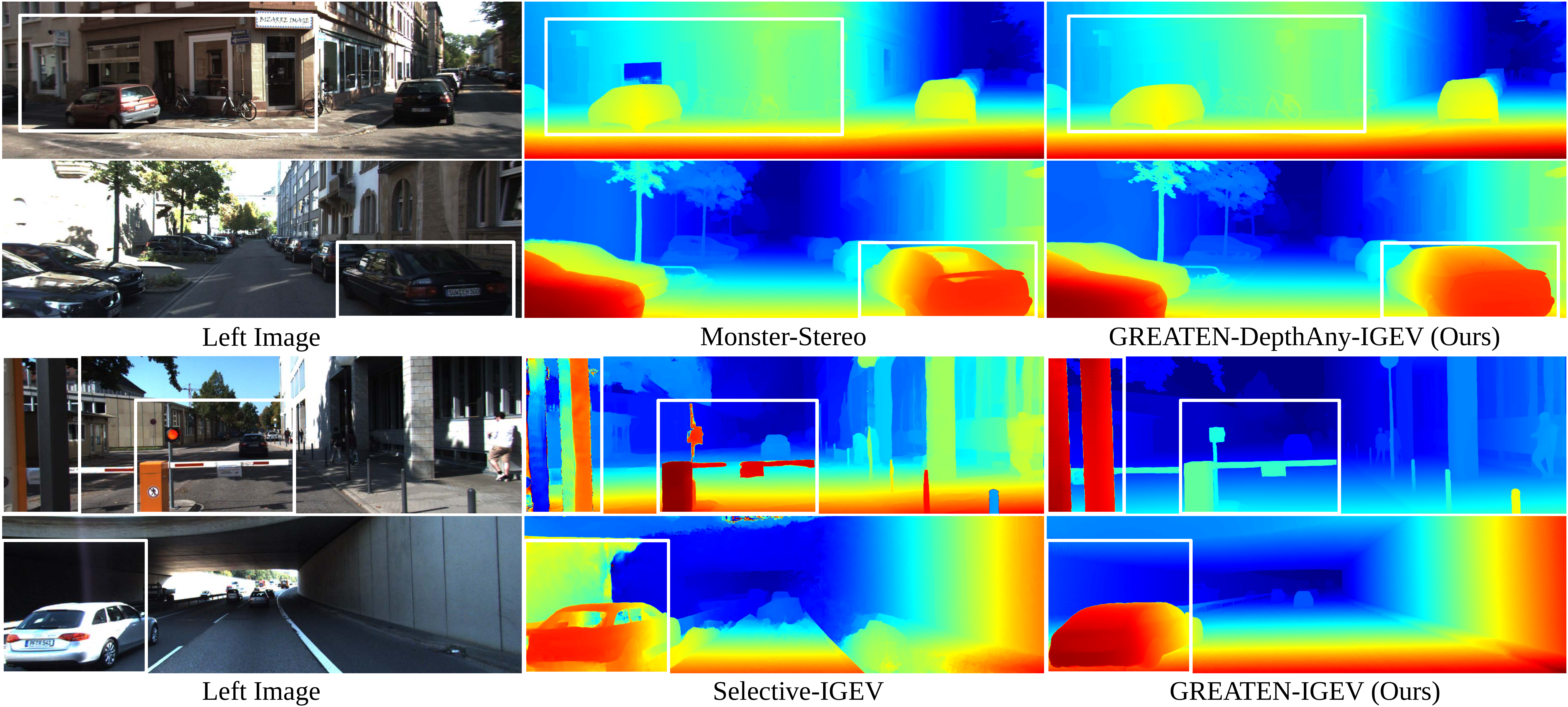}
    \caption{Zero-Shot qualitative results on KITTI testing set. \textbf{Row 1 \& 2} are the KITTI-2012 \cite{kitti2012}, and \textbf{Row 3 \& 4} are the KITTI-2015 \cite{kitti2015}. Our GREATEN-IGEV and GREATEN-DepthAny-IGEV outperform other iterative methods, producing clearer and more complete geometric structures. All the models are trained exclusively on Synthetic datasets.}
    \label{fig:kitti_zero_shot_qualitative_results}
\end{figure}

\begin{table*}[!t]
    \caption{In-Domain Performance on SceneFlow \cite{sceneflow} test set.}
    \label{tab:sceneflow_results}
    \centering
    \resizebox{2.0\columnwidth}{!}{
        \begin{tabular}{l|cccccccccc}
            \toprule
            Metrics & GOAT-Stereo \cite{goatstereo} & RAFT-Stereo \cite{raftstereo} & IGEV++ \cite{igev++} & IGEV-Stereo \cite{igevstereo} & Selective-IGEV \cite{selectivestereo} & GREAT-IGEV \cite{greatstereo} & GREATEN-IGEV (Ours) \\
            \hline
            EPE-Noc $ \downarrow $ & - & 0.21 & - & 0.19 & \trd 0.17 & \snd 0.14 & \fst \textbf{0.12} \\
            EPE-Occ $ \downarrow $ & \snd 1.53 & 1.94 & - & 1.65 & \trd 1.57 & \fst \textbf{1.51} & \fst \textbf{1.51} \\
            EPE-All $ \downarrow $ & 0.47 & 0.55 & \trd 0.43 & 0.48 & 0.45 & \snd 0.41 & \fst \textbf{0.39} \\
            \bottomrule
        \end{tabular}
    }
\end{table*}

\begin{table*}[!t]
    \caption{In-Domain Performance on Robust Vision Challenge (RVC).}
    \label{tab:rvc_results}
    \centering
    \resizebox{2.0\columnwidth}{!}{
        \begin{tabular}{l|cccccc|cccc}
            \toprule
            \multirow{3}{*}{Model} & \multicolumn{6}{c|}{KITTI-2015 \cite{kitti2015}} & \multicolumn{4}{c}{ETH3D \cite{eth3d}} \\
            \cmidrule{2-11}
            & \multicolumn{3}{c|}{All} & \multicolumn{3}{c|}{Noc} & \multicolumn{2}{c|}{All} & \multicolumn{2}{c}{Noc} \\
            & D1-bg $ \downarrow $ & D1-fg $ \downarrow $ & \multicolumn{1}{c|}{D1-all $ \downarrow $} & D1-bg $ \downarrow $ & D1-fg $ \downarrow $ & D1-all $ \downarrow $ & D1 $ \downarrow $ & \multicolumn{1}{c|}{EPE $ \downarrow $} & D1 $ \downarrow $ & EPE $ \downarrow $ \\
            \hline
            \multicolumn{11}{c}{\textcolor{magenta}{Robust Vision Challenge 2018}} \\
            \hline
            DN-CSS\_ROB \cite{dncssrob} & 2.39 & 5.71 & 2.94 & 2.23 & 4.96 & 2.68 & 3.00 & 0.24 & 2.69 & 0.22 \\
            iResNet\_ROB \cite{iresnetrob} & 2.27 & 4.89 & 2.71 & 2.10 & 3.96 & 2.40 & 4.67 & 0.27 & 4.23 & 0.25 \\
            \hline
            \multicolumn{11}{c}{\textcolor{magenta}{Robust Vision Challenge 2020}} \\
            \hline
            NLCA\_NET\_v2\_RVC \cite{nlcanet} & 1.51 & 3.97 & 1.92 & \trd 1.36 & 3.49 & 1.71 & 4.11 & 0.29 & 3.84 & 0.27 \\
            CFNet\_RVC \cite{cfnet} & 1.65 & 3.53 & 1.96 & 1.50 & 3.03 & 1.76 & 3.70 & 0.26 & 3.31 & 0.24 \\
            \hline
            \multicolumn{11}{c}{\textcolor{magenta}{Robust Vision Challenge 2022}} \\
            \hline
            iRaftStereo\_RVC \cite{improvedraftstereo, raftstereo} & 1.88 & 3.03 & 2.07 & 1.76 & 2.94 & 1.95 & 1.88 & \trd 0.17 & 1.62 & \trd 0.16 \\
            CREStereo++\_RVC \cite{crestereo++} & 1.55 & 3.53 & 1.88 & 1.43 & 3.36 & 1.75 & 1.70 & \snd 0.16 & 1.59 & \snd 0.15 \\
            \hline
            \multicolumn{11}{c}{\textcolor{magenta}{CVPR 2025 \& ICCV 2025}} \\
            \hline
            DEFOM-Stereo\_RVC \cite{defomstereo} & \snd 1.42 & \fst \textbf{2.68} & \snd 1.63 & \snd 1.32 & \fst 2.67 & \snd 1.54 & \snd 1.09 & \fst \textbf{0.13} & \trd 0.98 & \fst 0.13 \\
            SMoE-Stereo\_RVC \cite{smoestereo} & \trd 1.50 & \trd 2.88 & \trd 1.73 & 1.40 & \trd 2.86 & \trd 1.64 & \trd 1.13 & \fst 0.14 & \snd 0.95 & \fst 0.13 \\
            \hline
            GREATEN-IGEV\_RVC (Ours) & \fst \textbf{1.28} & \snd 2.86 & \fst \textbf{1.54} & \fst \textbf{1.19} & \snd 2.76 & \fst \textbf{1.45} & \fst 0.74 & \fst 0.14 & \fst \textbf{0.50} & \fst \textbf{0.12} \\
            GREATEN-DepthAny-IGEV\_RVC (Ours) & \fst 1.36 & \fst 2.82 & \fst 1.61 & \fst 1.26 & \fst \textbf{2.52} & \fst 1.47 & \fst \textbf{0.73} & \fst 0.14 & \fst 0.53 & \fst 0.13 \\
            \bottomrule
        \end{tabular}
    }
\end{table*}

\subsection{Comparisons with Existing Representative Methods}
\label{sec:comparisons_with_sota}

\begin{figure*}[tp]
    \centering
    \includegraphics[width=0.97\linewidth]{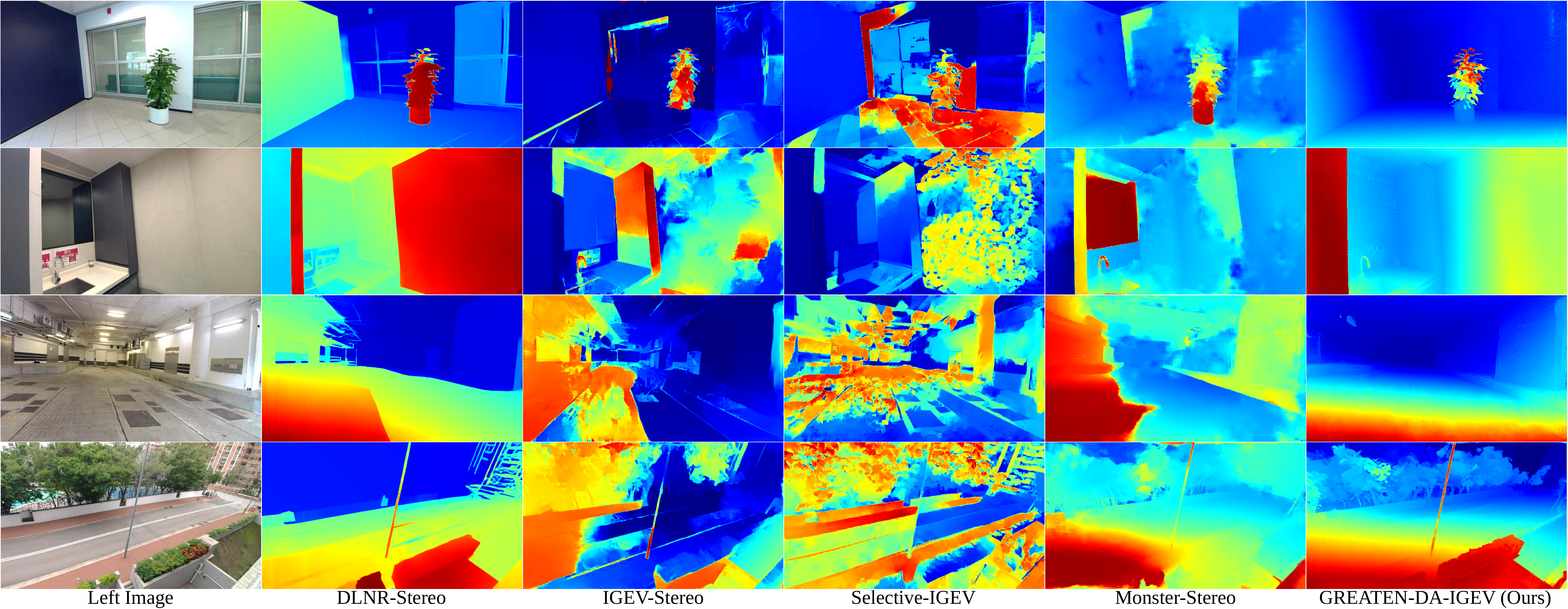}
    \caption{Zero-Shot qualitative results on our captured real-world data. Our GREATEN-DepthAny-IGEV outperforms other iterative methods, where "DA" stands for "DepthAny".  All the models are trained exclusively on Synthetic datasets.}
    \label{fig:in_the_wild_zero_shot_qualitative_results}
\end{figure*}

\textbf{Syn-to-Real Generalization on KITTI and Booster.} Trained exclusively on the synthetic SceneFlow dataset \cite{sceneflow}, GREATEN-Stereo exhibits outstanding Syn-to-Real generalization on the KITTI \cite{kitti2012, kitti2015} and Booster \cite{booster} benchmarks, as reported in Tab. \ref{tab:zero_shot_for_kitti_and_booster}. Among the VFM-Free variants, GREATEN-IGEV and GREATEN-Selective achieve D3-Noc metrics of 2.27 and 4.12 on KITTI-2012 \cite{kitti2012} and KITTI-2015 \cite{kitti2015}, respectively, corresponding to relative reductions of 26.3\% over IGEV-Stereo \cite{igevstereo} and 13.3\% over Selective-IGEV \cite{selectivestereo}. On the challenging non-Lambertian Booster benchmark \cite{booster}, GREATEN-IGEV attains an EPE of 2.85, outperforming RAFT-Stereo \cite{raftstereo} by 20.6\%. Notably, this result also surpasses several specialized VFM-Enhanced models, yielding EPE reductions of 8.9\% relative to Bridge-Depth \cite{bridgedepth} and 6.3\% relative to Monster-Stereo \cite{monsterstereo}. For the VFM-Enhanced variant, GREATEN-DepthAny-IGEV improves the D4 metric on Booster \cite{booster} from 9.53 achieved by DEFOM-Stereo \cite{defomstereo} to 8.3, marking a 12.9\% reduction, and advances the D6 metric from 7.54 achieved by Prompt-Stereo \cite{promptstereo} to 6.32, reflecting a 16.2\% reduction. Furthermore, after training on our proposed mixed Syn-to-Real datasets, both the VFM-Free and VFM-Enhanced GREATEN variants achieve state-of-the-art performance on all KITTI \cite{kitti2012, kitti2015} and Booster \cite{booster} metrics. The qualitative results in Fig. \ref{fig:booster_zero_shot_qualitative_results} and \ref{fig:kitti_zero_shot_qualitative_results} further corroborate the superior Syn-to-Real generalization of GREATEN-Stereo. In particular, the predictions exhibit clearer and more complete geometric structures for transparent objects, specular reflections, and large textureless regions. Meanwhile, as shown in Fig. \ref{fig:in_the_wild_zero_shot_qualitative_results}, our GREATEN-DepthAny-IGEV demonstrates splendid zero-shot performance on our captured realistic data, significantly outperforming existing stereo matching methods that largely fail under such conditions.

\textbf{Syn-to-Real Generalization on ETH3D and Middlebury.} As reported in Tab. \ref{tab:zero_shot_for_eth3d_and_middlebury}, the proposed GREATEN framework demonstrates superior Syn-to-Real generalization on ETH3D \cite{eth3d} and Middlebury-Half \cite{middlebury}. Specifically, our VFM-Free variant, GREATEN-IGEV, improves D1-Noc on ETH3D \cite{eth3d} by 30\% relative to FoundationStereo \cite{foundationstereo} and by 55.9\% relative to Monster-Stereo \cite{monsterstereo}. Meanwhile, the VFM-Enhanced variant, GREATEN-DepthAny-IGEV, achieves an EPE of 0.81 on Middlebury-Half \cite{middlebury}, representing a 37.2\% improvement over SMoE-Stereo \cite{smoestereo}. Furthermore, after training on our proposed mixed Syn-to-Real datasets, both the VFM-Free and VFM-Enhanced GREATEN variants achieve state-of-the-art performance across all ETH3D \cite{eth3d} and Middlebury \cite{middlebury} metrics. The qualitative results in Fig. \ref{fig:eth3d_middlebury_zero_shot_qualitative_results} further show that GREATEN-Stereo remains robust in challenging cases, including complex occluded regions (Row 1) and transparent surfaces (Row 3).

\begin{figure*}[tp]
    \centering
    \includegraphics[width=0.97\linewidth]{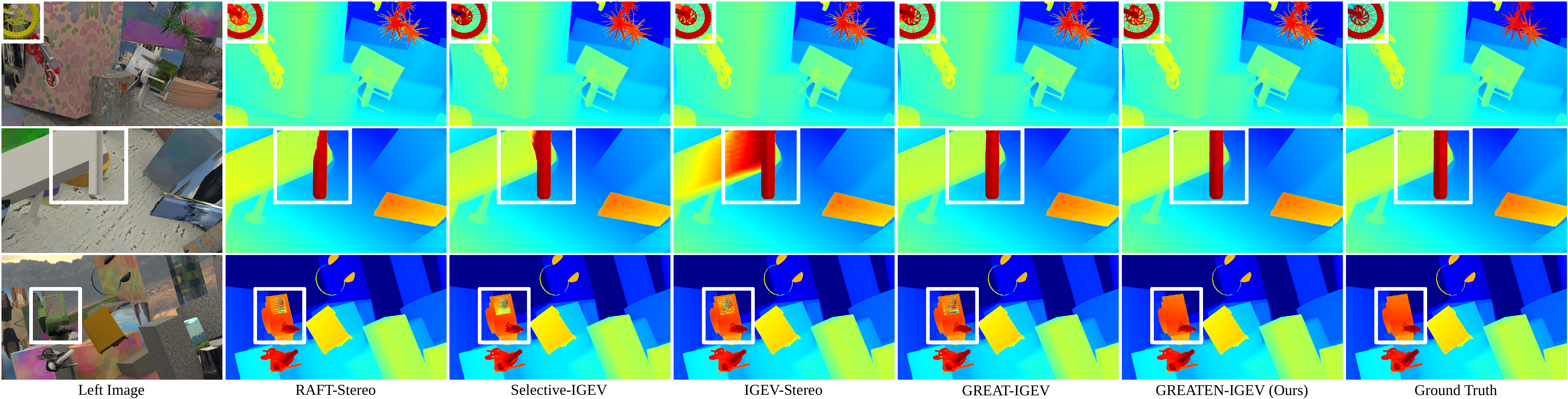}
    \caption{In-Domain qualitative results on SceneFlow \cite{sceneflow} test set. Our GREATEN-IGEV outperforms other iterative methods in occluded (\textbf{Row 1}), textureless (\textbf{Row 2}), and repetitive texture (\textbf{Row 3}) regions.}
    \label{fig:sceneflow_qualitative_results}
\end{figure*}

\textbf{In-Domain Performance on SceneFlow.} In addition to Syn-to-Real generalization, we further evaluate the in-domain performance of GREATEN-Stereo on the synthetic SceneFlow \cite{sceneflow} test set. As shown in Tab. \ref{tab:sceneflow_results}, GREATEN-IGEV ranks first on the SceneFlow \cite{sceneflow} test set in terms of EPE across all regions, encompassing both occluded and non-occluded areas. Specifically, compared with IGEV++ \cite{igev++} and IGEV-Stereo \cite{igevstereo}, GREATEN-IGEV reduces EPE-All and EPE-Noc by 9.3\% and 36.8\%, respectively. Despite replacing all designs in GREAT-IGEV \cite{greatstereo} with sparse counterparts, GREATEN-IGEV still preserves the EPE-Occ performance on the SceneFlow \cite{sceneflow} test set, attaining a score of 1.51 that remains on par with GREAT-IGEV \cite{greatstereo}. This performance retention mainly stems from the inherent filtering capability of Sparse Attentions, which suppresses noise introduced by redundant information in dense attention mechanisms and provides greater flexibility in modeling global irregular object contexts. The qualitative results in Fig. \ref{fig:sceneflow_qualitative_results} further show that GREATEN-IGEV produces more discriminative geometric structures in cases where GREAT-IGEV fails (Row 1 \& 3).

\textbf{In-Domain Performance on Robust Vision Challenge.} Beyond Syn-to-Real generalization, we further verify the in-domain real-world performance of GREATEN-Stereo on the KITTI-2015 \cite{kitti2015} and ETH3D \cite{eth3d} leaderboards through the Robust Vision Challenge (RVC). As shown in Tab. \ref{tab:rvc_results}, under the RVC training pipeline, GREATEN-IGEV and GREATEN-DepthAny-IGEV rank first across both benchmarks. On KITTI-2015 \cite{kitti2015}, GREATEN-IGEV achieves D1-All-all and D1-Noc-all errors of 1.54 and 1.45, respectively, yielding improvements of 5.5\% and 5.8\% over DEFOM-Stereo \cite{defomstereo}. On ETH3D \cite{eth3d}, GREATEN-DepthAny-IGEV obtains D1-All and D1-Noc errors of 0.73 and 0.53, respectively, surpassing SMoE-Stereo \cite{smoestereo} by 35.4\% and 44.2\%.

\subsection{Framework Universality and Efficiency}
\label{sec:framework_universality_and_efficiency}

\textbf{Framework Universality.} To evaluate the universality of the proposed GREATEN framework, we integrate it with two baselines, IGEV-Stereo \cite{igevstereo} and Selective-IGEV \cite{selectivestereo}, resulting in GREATEN-IGEV and GREATEN-Selective, respectively. As shown in Tab. \ref{tab:zero_shot_for_kitti_and_booster} and \ref{tab:zero_shot_for_eth3d_and_middlebury}, in Syn-to-Real generalization settings, GREATEN-IGEV and GREATEN-Selective reduce the EPE on Middlebury \cite{middlebury} and ETH3D \cite{eth3d} by 66.2\% and 83.8\%, respectively, compared to their corresponding baselines. Additionally, when trained solely on SceneFlow \cite{sceneflow}, GREATEN-IGEV surpasses GREAT-IGEV \cite{greatstereo} by 38.2\% in terms of EPE on Booster \cite{booster}. These results indicate that the GREATEN framework can be seamlessly integrated into existing iterative stereo matching methods, delivering significant improvements in Syn-to-Real generalization. Furthermore, GREATEN substantially outperforms previous GREAT \cite{greatstereo}, particularly on the challenging non-Lambertian Booster dataset \cite{booster}.

\textbf{Computational Overhead.} To reduce the computational cost of the GREAT framework \cite{greatstereo}, we replace all of its components with sparse alternatives. As reported in Tab. \ref{tab:computational_overhead} (a), GREATEN-IGEV achieves substantially lower inference times than GREAT-IGEV across all three datasets (SceneFlow \cite{sceneflow}, KITTI-2015 \cite{kitti2015}, and Middlebury \cite{middlebury}). Specifically, on Middlebury-Half \cite{middlebury} with a maximum disparity of 192 and an image resolution of 970$ \times $1470, GREATEN-IGEV reduces inference time by 19.2\% compared to GREAT-IGEV \cite{greatstereo}. On KITTI-2015 \cite{kitti2015}, with the maximum disparity set to 192 and the image resolution fixed at 375$ \times $1242, GREATEN-IGEV reduces inference time by 16.7\% compared to GREAT-IGEV \cite{greatstereo}. In addition, our VFM-Enhanced GREATEN-DepthAny-IGEV also achieves lower inference times, outperforming Monster-Stereo \cite{monsterstereo} by 5.1\% on SceneFlow \cite{sceneflow} with the maximum disparity set to 192 and an image resolution of 540$ \times $960. Furthermore, enabled by these sparse alternatives, GREATEN-IGEV successfully performs inference on full-resolution Middlebury \cite{middlebury} images with the maximum disparity increased to 768 and image resolutions close to 3K. In contrast, due to its full-attention mechanisms, GREAT-IGEV \cite{greatstereo} encounters out-of-memory errors under identical settings, even on an NVIDIA A800-80G GPU, as shown in Tab. \ref{tab:computational_overhead} (b).

\begin{figure}[tp]
    \centering
    \includegraphics[width=0.97\linewidth]{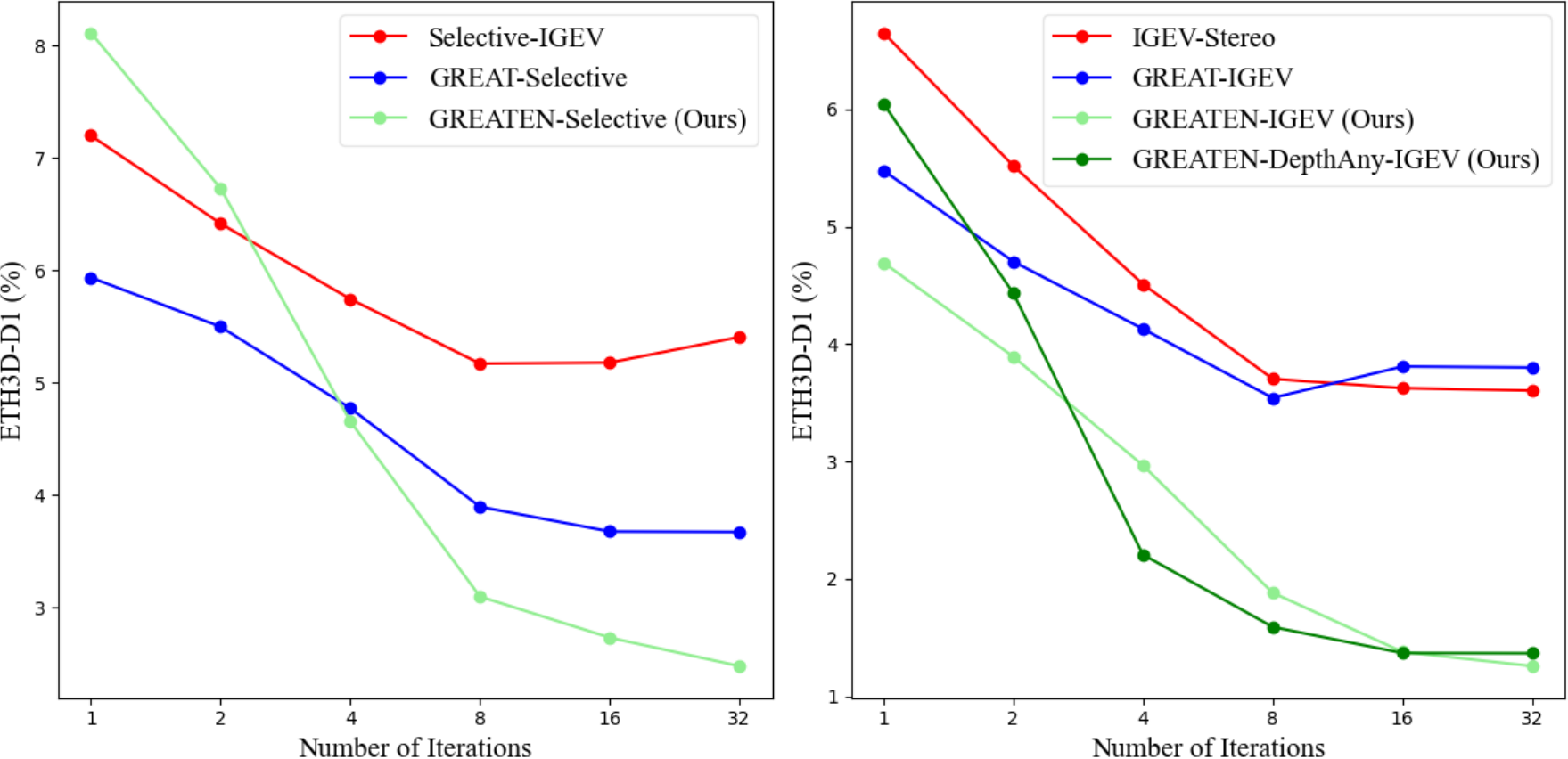}
    \caption{Convergence of the number of iterations. Results report the D1-Noc for Syn-to-Real generalization on the ETH3D \cite{eth3d}. The proposed GREATEN framework demonstrates faster convergence.}
    \label{fig:num_of_iterations}
\end{figure}

\textbf{Number of Iterations.} As illustrated in Fig. \ref{fig:num_of_iterations}, the proposed GREATEN framework substantially accelerates the convergence of the Convolutional Gated Recurrent Unit (ConvGRU) under the D1-Noc metric during Syn-to-Real generalization on ETH3D \cite{eth3d}. Meanwhile, GREATEN-Stereo maintains a stable convergence trend compared to Selective-IGEV \cite{selectivestereo} and GREAT-IGEV \cite{greatstereo}, whose D1-Noc errors on ETH3D increase after 8 iterations. Specifically, as reported in Tab. \ref{tab:num_of_iterations}, GREATEN-Selective requires only 8 iterations to outperform the fully inferred Selective-IGEV \cite{selectivestereo} and GREAT-Selective \cite{greatstereo}. Moreover, GREATEN-IGEV surpasses both the fully inferred IGEV-Stereo \cite{igevstereo} and GREAT-IGEV \cite{greatstereo} with only 4 iterations, achieving 8$ \times $ faster convergence.

\begin{table*}[!t]
    \caption{Comparison of the computational overhead. "DA" denotes "DepthAny". Results in Table (a) are evaluated on NVIDIA RTX 4090. Results in Table (b) are inferred on NVIDIA A800-80GB using Full-Resolution Middlebury and max disparity set to 768.}
    \label{tab:computational_overhead}
    \centering
    \resizebox{1.12\columnwidth}{!}{
        \begin{tabular}{l|c|c|c}
            \toprule
            Model (iterations: 32) & SceneFlow \cite{sceneflow} & KITTI-2015 \cite{kitti2015} & Midd (H) \cite{middlebury} \\
            \hline
            \multicolumn{4}{c}{VFM-Free} \\
            \hline
            IGEV-Stereo \cite{igevstereo} & 0.18s & 0.17s & 0.56s \\
            \rowcolor{thirdcolor} GREAT-IGEV \cite{greatstereo} & 0.27s & 0.24s & 0.73s \\
            \rowcolor{firstcolor} GREATEN-IGEV (Ours) & 0.22s & 0.20s & 0.59s \\
            \hline
            \multicolumn{4}{c}{VFM-Enhanced} \\
            \hline
            Monster-Stereo \cite{monsterstereo} & 0.39s & 0.35s & 1.45s \\
            GREATEN-DA-IGEV (Ours) & 0.37s & 0.34s & 1.16s \\
            \bottomrule
            \multicolumn{4}{c}{(a) Comparison of Inference Time ("Midd" for Middlebury).} \\
        \end{tabular}
    }
    \resizebox{0.88\columnwidth}{!}{
        \begin{tabular}{l|c|c}
            \toprule
            Model (iterations: 32) & Memory (GB) & Param (MB) \\
            \hline
            Selective-IGEV \cite{selectivestereo} & 28.9 & 13.1 \\
            \rowcolor{thirdcolor} GREAT-Selective \cite{greatstereo} & Out-of-Memory & 15.0 \\
            \rowcolor{firstcolor} GREATEN-Selective & 48.4 & 22.2 \\
            \hline
            IGEV-Stereo \cite{igevstereo} & 23.1 & 12.6 \\
            \rowcolor{thirdcolor} GREAT-IGEV \cite{greatstereo} & Out-of-Memory & 14.4 \\
            \rowcolor{firstcolor} GREATEN-IGEV (Ours) & 45.9 & 21.7 \\
            GREATEN-DA-IGEV (Ours) & 47.8 & 359.4 \\
            \bottomrule
            \multicolumn{3}{c}{(b) Comparison of CUDA Memory Consumption.} \\
        \end{tabular}
    }
\end{table*}

\subsection{Ablation Study}
\label{sec:ablation_study}

\textbf{Effectiveness of Sparse Attentions.} To evaluate the sparse attention modules in the proposed framework, we adopt GREAT-IGEV \cite{greatstereo} as the baseline and replace its corresponding components with Sparse Spatial Attention (SSA), Sparse Dual-Matching Attention (SDMA), and Simple Volume Attention (SVA). To exclude the influence of surface normals from all sparse attention variants, the surface normal branch within SDMA is disabled. As shown in Tab. \ref{tab:ablation_study_of_sparse_alternatives}, SVA maintains the overall performance of the full-attention-based Volume Attention (VA) in GREAT-IGEV \cite{greatstereo}. Moreover, despite utilizing a substantially simplified 4D volume filter, SVA improves EPE-Occ and D3-Occ on the SceneFlow \cite{sceneflow} test set from 1.514 and 10.115 in GREAT-IGEV \cite{greatstereo} to 1.487 and 9.86, corresponding to reductions of 1.8\% and 2.5\%, respectively. Meanwhile, SSA and SDMA (with only the image-context branch) yield further overall improvements across all metrics on the SceneFlow \cite{sceneflow} test set. As further reported in Tab. \ref{tab:ablation_study_of_sparse_alternatives}, when all sparse alternatives are utilized, D3-All, D3-Noc, and D3-Occ on the SceneFlow \cite{sceneflow} test set decrease from 2.201, 0.486, and 10.115 to 2.122, 0.468, and 9.014, respectively, representing relative reductions of 3.6\%, 3.7\%, and 10.9\% compared to GREAT-IGEV \cite{greatstereo}. Benefiting from Key Points Sampling with its flexible adaptation to object contexts and the ability to suppress redundant details, these sparse attentions can model complex object contexts more precisely, thereby yielding a large improvement margin in occlusion-induced ill-posed regions (Fig. \ref{fig:sceneflow_qualitative_results}, Row 1).

\begin{table}[!t]
    \caption{Comparison of the number of iterations. "DA" denotes "DepthAny". Results report the D1-Noc $ \downarrow $ for Syn-to-Real generalization on the ETH3D \cite{eth3d}}
    \label{tab:num_of_iterations}
    \centering
    \resizebox{1.0\columnwidth}{!}{
        \begin{tabular}{lcccccc}
            \toprule
            \multirow{2}{*}{Model} & \multicolumn{6}{c}{Number of Iterations} \\
            \cmidrule{2-7}
            & 1 & 2 & 4 & 8 & 16 & 32 \\
            \hline
            Selective-IGEV \cite{selectivestereo} & 7.20 & 6.42 & 5.74 & 5.17 & 5.18 & 5.41 \\
            \rowcolor{thirdcolor} GREAT-Selective \cite{greatstereo} & 5.94 & 5.50 & 4.77 & 3.90 & 3.68 & 3.67 \\
            \rowcolor{firstcolor} GREATEN-Selective (Ours) & 8.11 & 6.73 & 4.66 & 3.10 & 2.73 & 2.48 \\
            \hline
            IGEV-Stereo \cite{igevstereo} & 6.65 & 5.52 & 4.51 & 3.70 & 3.63 & 3.60 \\
            \rowcolor{thirdcolor} GREAT-IGEV \cite{greatstereo} & 5.48 & 4.70 & 4.13 & 3.54 & 3.81 & 3.80 \\
            \rowcolor{firstcolor} GREATEN-IGEV (Ours) & 4.69 & 3.89 & 2.97 & 1.88 & 1.38 & 1.26 \\
            GREATEN-DA-IGEV (Ours) & 6.05 & 4.43 & 2.21 & 1.59 & 1.37 & 1.37 \\
            \bottomrule
        \end{tabular}
    }
\end{table}

\textbf{Effectiveness of Surface Normals Guided Modules.} The effectiveness of Sparse Attentions in capturing object textural context is highly dependent on image appearance. Consequently, image-domain shifts degrade their Syn-to-Real generalization, particularly in challenging non-Lambertian regions. As reported in Tab. \ref{tab:ablation_study_of_sta_gcgf}, when all Sparse Attention modules are applied, the D2 metric on the non-Lambertian Booster \cite{booster} benchmark increases from 19.3 in GREAT-IGEV \cite{greatstereo} to 25.29, corresponding to a 31\% performance drop. To alleviate this issue, we introduce a Gated Contextual-Geometric Fusion (GCGF) module that incorporates surface normals. Compared to the variant using all Sparse Attention modules alone, GCGF improves overall Syn-to-Real generalization. Specifically, under the Syn-to-Real setting, GCGF reduces the D1 error on ETH3D \cite{eth3d} from 3.71 to 1.71 and the D2 error on Booster \cite{booster} from 25.29 to 19.53, yielding relative reductions of 53.9\% and 22.8\%, respectively. However, as illustrated in Fig. \ref{fig:specular_transparent_augmentation_affect}, the simplified illumination and limited noise characteristics of the synthetic SceneFlow \cite{sceneflow} dataset restrict the effectiveness of GCGF. To address this limitation, we further introduce Specular-Transparent Augmentation (STA). As shown in Tab. \ref{tab:ablation_study_of_sta_gcgf}, the full GREATEN-IGEV model with STA achieves further improvements in overall Syn-to-Real generalization across all metrics on ETH3D \cite{eth3d}, Booster \cite{booster}, KITTI-2012 \cite{kitti2012}, and KITTI-2015 \cite{kitti2015}, compared to the baseline GREAT-IGEV \cite{greatstereo}. In particular, GREATEN-IGEV reduces D3-Noc on KITTI-2012 \cite{kitti2012} from 3.35 to 2.27 and D4 on Booster \cite{booster} from 14.6 to 11.1, corresponding to relative improvements of 32.2\% and 24\%, respectively. Furthermore, inspired by the side-feature design in FoundationStereo \cite{foundationstereo} and the learnable metric-depth alignment in DEFOM-Stereo \cite{defomstereo}, we develop a VFM-Enhanced variant, termed GREATEN-DepthAny-IGEV. As shown in Tab. \ref{tab:ablation_study_of_sta_gcgf}, this variant further improves Syn-to-Real generalization over the baseline GREAT-IGEV \cite{greatstereo}.

\begin{table*}[!t]
    \caption{Ablation study of the effectiveness for Simple Volume Attention (SVA), Sparse Spatial Attention (SSA), and Sparse Dual-Matching Attention (SDMA) on the SceneFlow \cite{sceneflow} test set (\textbf{bold}: best). For the following results, the surface normal branch in SDMA is disabled to isolate the surface-normal-related modules. Models are trained on SceneFlow \cite{sceneflow}.}
    \label{tab:ablation_study_of_sparse_alternatives}
    \centering
    \begin{tabular}{l|ccc|cc|cc|cc|c}
        \toprule
        \multirow{2}{*}{Model} & \multirow{2}{*}{SVA} & \multirow{2}{*}{SSA} & \multirow{2}{*}{SDMA} & \multicolumn{2}{c|}{All} & \multicolumn{2}{c|}{Noc} & \multicolumn{2}{c|}{Occ} & 
        \multirow{2}{*}{Params (MB)} \\
        \cmidrule{5-10}
        & & & & EPE $ \downarrow $ & D3 $ \downarrow $ & EPE $ \downarrow $ & D3 $ \downarrow $ & EPE $ \downarrow $ & D3 $ \downarrow $ & \\
        \hline
        IGEV-Stereo \cite{igevstereo} & & & & 0.479 & 2.476 & 0.194 & 0.633 & 1.649 & 10.980 & 12.60 \\
        Baseline (GREAT-IGEV) \cite{greatstereo} & & & & 0.413 & 2.201 & 0.135 & 0.486 & 1.514 & 10.115 & 14.44 \\
        \hline
        SVA & \checkmark & & & 0.415 & 2.158 & 0.152 & 0.489 & 1.487 & 9.860 & 14.41 \\
        SVA + SSA & \checkmark & \checkmark & & 0.390 & 2.133 & 0.118 & 0.473 & 1.485 & 9.080 & 15.10 \\
        SVA + SSA + SDMA & \checkmark & \checkmark & \checkmark & \textbf{0.382} & \textbf{2.122} & \textbf{0.115} & \textbf{0.468} & \textbf{1.469} & \textbf{9.014} & 14.49 \\
        \bottomrule
    \end{tabular}
\end{table*}

\begin{table*}[!t]
    \caption{Ablation study of the effectiveness for Gated Contextual-Geometric Fusion (GCGF) and Specular-Transparent Augmentation (STA) on the Syn-to-Real generalization across four real-world benchmarks (ETH3D \cite{eth3d}, Booster \cite{booster}, KITTI-2012 \cite{kitti2012}, and KITTI-2015 \cite{kitti2015}). \textbf{Bold} for the best and \underline{underline} for the second. "DA" denotes "DepthAny". For the following results, the surface normal branch in SDMA is blocked when GCGF is disabled. Models are trained on SceneFlow \cite{sceneflow}.}
    \label{tab:ablation_study_of_sta_gcgf}
    \centering
    \resizebox{2.0\columnwidth}{!}{
        \begin{tabular}{l|ccc|c|cc|ccc|ccc|c}
            \toprule
            \multirow{3}{*}{Model} & \multirow{3}{*}{GCGF} & \multirow{3}{*}{STA} & \multirow{3}{*}{VFM-Enhanced} & \multirow{2}{*}{ETH3D \cite{eth3d}} & \multicolumn{2}{c|}{\multirow{2}{*}{Booster \cite{booster}}} & \multicolumn{3}{c|}{KITTI-2012 \cite{kitti2012}} & \multicolumn{3}{c|}{KITTI-2015 \cite{kitti2015}} & \multirow{3}{*}{Params (MB)} \\
            \cmidrule{8-13}
            & & & & & & & \multicolumn{3}{c|}{D3 $ \downarrow $} & \multicolumn{3}{c|}{D3 $ \downarrow $} & \\
            \cmidrule{5-13}
            & & & & D1 $ \downarrow $ & D2 $ \downarrow $ & D4 $ \downarrow $ & All & Occ & Noc & All & Occ & Noc & \\
            \hline
            IGEV-Stereo \cite{igevstereo} & & & & 3.61 & \underline{17.47} & 13.79 & 5.13 & 6.28 & 3.08 & 6.03 & 6.69 & 4.56 & 12.60 \\
            Baseline (GREAT-IGEV) \cite{greatstereo} & & & & 3.80 & 19.30 & 14.60 & 5.31 & 6.40 & 3.35 & 5.88 & 6.53 & 4.41 & 14.44 \\
            \hline
            Sparse (SVA + SSA + SDMA) & & & & 3.71 & 25.29 & 17.83 & 6.96 & 7.88 & 5.72 & 6.40 & 7.02 & 5.01 & 14.49 \\
            \hline
            Sparse + GCGF & \checkmark & & & 1.71 & 19.53 & 12.54 & 5.75 & 6.88 & 4.05 & 5.84 & 6.60 & 4.15 & 21.68 \\
            Full Model (GREATEN-IGEV) & \checkmark & \checkmark & & \textbf{1.26} & 17.55 & \underline{11.10} & \underline{4.54} & \underline{5.81} & \underline{2.27} & \underline{5.70} & \underline{6.41} & \underline{4.09} & 21.68 \\
            Full Model (GREATEN-DA-IGEV) & \checkmark & \checkmark & \checkmark & \underline{1.37} & \textbf{13.99} & \textbf{8.30} & \textbf{3.95} & \textbf{5.03} & \textbf{2.00} & \textbf{5.06} & \textbf{5.75} & \textbf{3.51} & 359.39 \\
            \bottomrule
        \end{tabular}
    }
\end{table*}

\begin{table}[!t]
    \caption{Ablation study on the effect of surface normal quality on the GREATEN framework (\textbf{Bold}: best). The model is trained and evaluated on SceneFlow \cite{sceneflow}.}
    \label{tab:ablation_study_of_quality_of_surface_normals}
    \centering
    \resizebox{1.0\columnwidth}{!}{
        \begin{tabular}{c|cc|cc|cc}
            \toprule
            Surface Normal & \multicolumn{2}{c|}{All} & \multicolumn{2}{c|}{Noc} & \multicolumn{2}{c}{Occ} \\
            \cmidrule{2-7}
            Generator & EPE $ \downarrow $ & D3 $ \downarrow $ & EPE $ \downarrow $ & D3 $ \downarrow $ & EPE $ \downarrow $ & D3 $ \downarrow $ \\
            \hline
            \multicolumn{7}{c}{GREATEN-IGEV} \\
            \hline
            ViT-Small & 0.439 & 2.394 & 0.138 & 0.573 & 1.673 & 10.064 \\
            ViT-Base & 0.408 & 2.252 & 0.127 & 0.512 & 1.563 & 9.574 \\
            ViT-Large & \textbf{0.394} & \textbf{2.196} & \textbf{0.123} & \textbf{0.494} & \textbf{1.514} & \textbf{9.374} \\
            \bottomrule
        \end{tabular}
    }
\end{table}

\textbf{Effect of Surface Normal Quality.} Since surface normals serve as one of the primary input modalities in the GREATEN framework, we further conduct experiments to examine how their quality affects the proposed framework. Specifically, following the pipeline of StereoAnywhere \cite{stereoanywhere}, we generate surface normals using three Vision Transformer (ViT) \cite{vit} variants with increasing model scales, namely ViT-Small, ViT-Base, and ViT-Large. As the ViT model size increases, the quality of the generated surface normals improves accordingly. As reported in Tab. \ref{tab:ablation_study_of_quality_of_surface_normals}, evaluations using GREATEN-IGEV on the SceneFlow \cite{sceneflow} test set demonstrate that the proposed GREATEN framework is sensitive to the quality of the input surface normals.

\section{Conclusion}
\label{sec:conclusion}

We propose GREATEN, a universal framework compatible with existing iterative stereo-matching methods. It is designed to improve Synthetic-to-Realistic Zero-Shot (Syn-to-Real) generalization by addressing the cross-domain shifts and ill-posed ambiguities inherent in image textures.

Specifically, GREATEN-Stereo incorporates a Gated Contextual-Geometric Fusion (GCGF) module to effectively integrate features from stereo images and surface normals. This module produces discriminative and domain-invariant contextual-geometric representations that enhance Syn-to-Real generalization. To further strengthen the robustness of GCGF against misleading visual cues, Specular-Transparent Augmentation (STA) is introduced to intentionally perturb the texture consistency of training images. In addition, by leveraging Sparse Spatial Attention (SSA), Sparse Dual-Matching Attention (SDMA), and Simple Volume Attention (SVA), GREATEN-Stereo substantially reduces the computational overhead compared to GREAT-Stereo.

When trained solely on the SceneFlow dataset, GREATEN-IGEV achieves strong Syn-to-Real generalization across five widely used real-world benchmarks: ETH3D, Middlebury, KITTI-2012, KITTI-2015, and Booster. Moreover, when enhanced by Vision Foundation Models, GREATEN-DepthAny-IGEV achieves state-of-the-art Syn-to-Real generalization across all five real-world datasets and demonstrates impressive generalization capability in our captured in-the-wild scenarios.

\section*{Acknowledgments}
The project is supported by a grant from the Hong Kong Research Grant Council under GRF 11219624.

\bibliographystyle{IEEEtran}
\bibliography{main_references}

@String(CVPR= {IEEE Conf. Comput. Vis. Pattern Recog.})

@String(ECCV= {Eur. Conf. Comput. Vis.})

@String(AAAI = {AAAI})

@String(CVPR  = {CVPR})

@String(ECCV  = {ECCV})

@inproceedings{sgbm,
  title={Accurate and efficient stereo processing by semi-global matching and mutual information},
  author={Hirschmuller, Heiko},
  booktitle={2005 IEEE computer society conference on computer vision and pattern recognition (CVPR'05)},
  volume={2},
  pages={807--814},
  year={2005},
  organization={IEEE}
}

@inproceedings{dispnet,
  title={A large dataset to train convolutional networks for disparity, optical flow, and scene flow estimation},
  author={Mayer, Nikolaus and Ilg, Eddy and Hausser, Philip and Fischer, Philipp and Cremers, Daniel and Dosovitskiy, Alexey and Brox, Thomas},
  booktitle={Proceedings of the IEEE conference on computer vision and pattern recognition},
  pages={4040--4048},
  year={2016}
}

@article{sssmnet,
  title={Self-supervised learning for stereo matching with self-improving ability},
  author={Zhong, Yiran and Dai, Yuchao and Li, Hongdong},
  journal={arXiv preprint arXiv:1709.00930},
  year={2017}
}

@inproceedings{gcnet,
  title={End-to-end learning of geometry and context for deep stereo regression},
  author={Kendall, Alex and Martirosyan, Hayk and Dasgupta, Saumitro and Henry, Peter and Kennedy, Ryan and Bachrach, Abraham and Bry, Adam},
  booktitle={Proceedings of the IEEE international conference on computer vision},
  pages={66--75},
  year={2017}
}

@inproceedings{dncssrob,
  title={Occlusions, motion and depth boundaries with a generic network for disparity, optical flow or scene flow estimation},
  author={Ilg, Eddy and Saikia, Tonmoy and Keuper, Margret and Brox, Thomas},
  booktitle={Proceedings of the European conference on computer vision (ECCV)},
  pages={614--630},
  year={2018}
}

@inproceedings{iresnetrob,
  title={Learning for disparity estimation through feature constancy},
  author={Liang, Zhengfa and Feng, Yiliu and Guo, Yulan and Liu, Hengzhu and Chen, Wei and Qiao, Linbo and Zhou, Li and Zhang, Jianfeng},
  booktitle={Proceedings of the IEEE conference on computer vision and pattern recognition},
  pages={2811--2820},
  year={2018}
}

@inproceedings{activenet,
  title={Activestereonet: End-to-end self-supervised learning for active stereo systems},
  author={Zhang, Yinda and Khamis, Sameh and Rhemann, Christoph and Valentin, Julien and Kowdle, Adarsh and Tankovich, Vladimir and Schoenberg, Michael and Izadi, Shahram and Funkhouser, Thomas and Fanello, Sean},
  booktitle={Proceedings of the european conference on computer vision (ECCV)},
  pages={784--801},
  year={2018}
}

@inproceedings{psmnet,
  title={Pyramid stereo matching network},
  author={Chang, Jia-Ren and Chen, Yong-Sheng},
  booktitle={Proceedings of the IEEE conference on computer vision and pattern recognition},
  pages={5410--5418},
  year={2018}
}

@inproceedings{gwcnet,
  title={Group-wise correlation stereo network},
  author={Guo, Xiaoyang and Yang, Kai and Yang, Wukui and Wang, Xiaogang and Li, Hongsheng},
  booktitle={Proceedings of the IEEE/CVF conference on computer vision and pattern recognition},
  pages={3273--3282},
  year={2019}
}

@inproceedings{ganet,
  title={Ga-net: Guided aggregation net for end-to-end stereo matching},
  author={Zhang, Feihu and Prisacariu, Victor and Yang, Ruigang and Torr, Philip HS},
  booktitle={Proceedings of the IEEE/CVF conference on computer vision and pattern recognition},
  pages={185--194},
  year={2019}
}

@inproceedings{aanet,
  title={Aanet: Adaptive aggregation network for efficient stereo matching},
  author={Xu, Haofei and Zhang, Juyong},
  booktitle={Proceedings of the IEEE/CVF conference on computer vision and pattern recognition},
  pages={1959--1968},
  year={2020}
}

@inproceedings{dsmnet,
  title={Domain-invariant stereo matching networks},
  author={Zhang, Feihu and Qi, Xiaojuan and Yang, Ruigang and Prisacariu, Victor and Wah, Benjamin and Torr, Philip},
  booktitle={European conference on computer vision},
  pages={420--439},
  year={2020},
  organization={Springer}
}

@inproceedings{cascademvsnet,
  title={Cascade cost volume for high-resolution multi-view stereo and stereo matching},
  author={Gu, Xiaodong and Fan, Zhiwen and Zhu, Siyu and Dai, Zuozhuo and Tan, Feitong and Tan, Ping},
  booktitle={Proceedings of the IEEE/CVF conference on computer vision and pattern recognition},
  pages={2495--2504},
  year={2020}
}

@inproceedings{cfnet,
  title={Cfnet: Cascade and fused cost volume for robust stereo matching},
  author={Shen, Zhelun and Dai, Yuchao and Rao, Zhibo},
  booktitle={Proceedings of the IEEE/CVF conference on computer vision and pattern recognition},
  pages={13906--13915},
  year={2021}
}

@article{pvnet,
  title={PVStereo: Pyramid voting module for end-to-end self-supervised stereo matching},
  author={Wang, Hengli and Fan, Rui and Cai, Peide and Liu, Ming},
  journal={IEEE Robotics and Automation Letters},
  volume={6},
  number={3},
  pages={4353--4360},
  year={2021},
  publisher={IEEE}
}

@article{sdanet,
  title={Synthetic-to-real domain adaptation joint spatial feature transform for stereo matching},
  author={Li, Xing and Fan, Yangyu and Rao, Zhibo and Lv, Guoyun and Liu, Shiya},
  journal={IEEE Signal Processing Letters},
  volume={29},
  pages={60--64},
  year={2021},
  publisher={IEEE}
}

@article{nlcanet,
  title={Rethinking training strategy in stereo matching},
  author={Rao, Zhibo and Dai, Yuchao and Shen, Zhelun and He, Renjie},
  journal={IEEE Transactions on Neural Networks and Learning Systems},
  volume={34},
  number={10},
  pages={7796--7809},
  year={2022},
  publisher={IEEE}
}

@inproceedings{pcwnet,
  title={Pcw-net: Pyramid combination and warping cost volume for stereo matching},
  author={Shen, Zhelun and Dai, Yuchao and Song, Xibin and Rao, Zhibo and Zhou, Dingfu and Zhang, Liangjun},
  booktitle={European conference on computer vision},
  pages={280--297},
  year={2022},
  organization={Springer}
}

@article{panet,
  title={Patch attention network with generative adversarial model for semi-supervised binocular disparity prediction},
  author={Rao, Zhibo and He, Mingyi and Dai, Yuchao and Shen, Zhelun},
  journal={The Visual Computer},
  volume={38},
  number={1},
  pages={77--93},
  year={2022},
  publisher={Springer}
}

@article{rsnet,
  title={Region separable stereo matching},
  author={Cheng, Junda and Yang, Xin and Pu, Yuechuan and Guo, Peng},
  journal={IEEE Transactions on Multimedia},
  volume={25},
  pages={4880--4893},
  year={2022},
  publisher={IEEE}
}

@article{uenet,
  title={Uncertainty estimation for stereo matching based on evidential deep learning},
  author={Wang, Chen and Wang, Xiang and Zhang, Jiawei and Zhang, Liang and Bai, Xiao and Ning, Xin and Zhou, Jun and Hancock, Edwin},
  journal={pattern recognition},
  volume={124},
  pages={108498},
  year={2022},
  publisher={Elsevier}
}

@inproceedings{graftnet,
  title={Graftnet: Towards domain generalized stereo matching with a broad-spectrum and task-oriented feature},
  author={Liu, Biyang and Yu, Huimin and Qi, Guodong},
  booktitle={Proceedings of the IEEE/CVF conference on computer vision and pattern recognition},
  pages={13012--13021},
  year={2022}
}

@inproceedings{acvnet,
  title={Attention concatenation volume for accurate and efficient stereo matching},
  author={Xu, Gangwei and Cheng, Junda and Guo, Peng and Yang, Xin},
  booktitle={Proceedings of the IEEE/CVF conference on computer vision and pattern recognition},
  pages={12981--12990},
  year={2022}
}

@article{fastacvnet,
  title={Accurate and efficient stereo matching via attention concatenation volume},
  author={Xu, Gangwei and Wang, Yun and Cheng, Junda and Tang, Jinhui and Yang, Xin},
  journal={IEEE Transactions on Pattern Analysis and Machine Intelligence},
  volume={46},
  number={4},
  pages={2461--2474},
  year={2023},
  publisher={IEEE}
}

@inproceedings{hvtnet,
  title={Domain generalized stereo matching via hierarchical visual transformation},
  author={Chang, Tianyu and Yang, Xun and Zhang, Tianzhu and Wang, Meng},
  booktitle={Proceedings of the IEEE/CVF Conference on Computer Vision and Pattern Recognition},
  pages={9559--9568},
  year={2023}
}

@inproceedings{maskednet,
  title={Masked representation learning for domain generalized stereo matching},
  author={Rao, Zhibo and Xiong, Bangshu and He, Mingyi and Dai, Yuchao and He, Renjie and Shen, Zhelun and Li, Xing},
  booktitle={Proceedings of the IEEE/CVF Conference on Computer Vision and Pattern Recognition},
  pages={5435--5444},
  year={2023}
}

@inproceedings{formernet,
  title={Learning representations from foundation models for domain generalized stereo matching},
  author={Zhang, Yongjian and Wang, Longguang and Li, Kunhong and Wang, Yun and Guo, Yulan},
  booktitle={European Conference on Computer Vision},
  pages={146--162},
  year={2024},
  organization={Springer}
}

@article{ercnet,
  title={Adaptively identify and refine ill-posed regions for accurate stereo matching},
  author={Liu, Changlin and Sun, Linjun and Ning, Xin and Xu, Jian and Yu, Lina and Zhang, Kaijie and Li, Weijun},
  journal={Neural Networks},
  volume={178},
  pages={106394},
  year={2024},
  publisher={Elsevier}
}

@article{coatrsnet,
  title={Coatrsnet: Fully exploiting convolution and attention for stereo matching by region separation},
  author={Cheng, Junda and Xu, Gangwei and Guo, Peng and Yang, Xin},
  journal={International Journal of Computer Vision},
  volume={132},
  number={1},
  pages={56--73},
  year={2024},
  publisher={Springer}
}

@article{dcanet,
  title={Cost volume aggregation in stereo matching revisited: A disparity classification perspective},
  author={Wang, Yun and Wang, Longguang and Li, Kunhong and Zhang, Yongjian and Wu, Dapeng Oliver and Guo, Yulan},
  journal={IEEE Transactions on Image Processing},
  volume={33},
  pages={6425--6438},
  year={2024},
  publisher={IEEE}
}

@article{adnet,
  title={Adstereo: Efficient stereo matching with adaptive downsampling and disparity alignment},
  author={Wang, Yun and Li, Kunhong and Wang, Longguang and Hu, Junjie and Wu, Dapeng Oliver and Guo, Yulan},
  journal={IEEE Transactions on Image Processing},
  year={2025},
  publisher={IEEE}
}

@inproceedings{dualnet,
  title={Dualnet: Robust self-supervised stereo matching with pseudo-label supervision},
  author={Wang, Yun and Zheng, Jiahao and Zhang, Chenghao and Zhang, Zhanjie and Li, Kunhong and Zhang, Yongjian and Hu, Junjie},
  booktitle={Proceedings of the AAAI Conference on Artificial Intelligence},
  volume={39},
  number={8},
  pages={8178--8186},
  year={2025}
}

@inproceedings{smoestereo,
  title={Learning robust stereo matching in the wild with selective mixture-of-experts},
  author={Wang, Yun and Wang, Longguang and Zhang, Chenghao and Zhang, Yongjian and Zhang, Zhanjie and Ma, Ao and Fan, Chenyou and Lam, Tin Lun and Hu, Junjie},
  booktitle={Proceedings of the IEEE/CVF International Conference on Computer Vision},
  pages={21276--21287},
  year={2025}
}

@inproceedings{raftstereo,
  title={Raft-stereo: Multilevel recurrent field transforms for stereo matching},
  author={Lipson, Lahav and Teed, Zachary and Deng, Jia},
  booktitle={2021 International conference on 3D vision (3DV)},
  pages={218--227},
  year={2021},
  organization={IEEE}
}

@article{improvedraftstereo,
  title={An improved raftstereo trained with a mixed dataset for the robust vision challenge 2022},
  author={Jiang, Hualie and Xu, Rui and Jiang, Wenjie},
  journal={arXiv preprint arXiv:2210.12785},
  year={2022}
}

@article{spstereo,
  title={SPNet: Learning stereo matching with slanted plane aggregation},
  author={Wang, Yun and Wang, Longguang and Wang, Hanyun and Guo, Yulan},
  journal={IEEE Robotics and Automation Letters},
  volume={7},
  number={3},
  pages={6258--6265},
  year={2022},
  publisher={IEEE}
}

@inproceedings{crestereo,
  title={Practical stereo matching via cascaded recurrent network with adaptive correlation},
  author={Li, Jiankun and Wang, Peisen and Xiong, Pengfei and Cai, Tao and Yan, Ziwei and Yang, Lei and Liu, Jiangyu and Fan, Haoqiang and Liu, Shuaicheng},
  booktitle={Proceedings of the IEEE/CVF conference on computer vision and pattern recognition},
  pages={16263--16272},
  year={2022}
}

@inproceedings{crestereo++,
  title={Uncertainty guided adaptive warping for robust and efficient stereo matching},
  author={Jing, Junpeng and Li, Jiankun and Xiong, Pengfei and Liu, Jiangyu and Liu, Shuaicheng and Guo, Yichen and Deng, Xin and Xu, Mai and Jiang, Lai and Sigal, Leonid},
  booktitle={Proceedings of the IEEE/CVF International Conference on Computer Vision},
  pages={3318--3327},
  year={2023}
}

@inproceedings{dlnrstereo,
  title={High-frequency stereo matching network},
  author={Zhao, Haoliang and Zhou, Huizhou and Zhang, Yongjun and Chen, Jie and Yang, Yitong and Zhao, Yong},
  booktitle={Proceedings of the IEEE/CVF conference on computer vision and pattern recognition},
  pages={1327--1336},
  year={2023}
}

@inproceedings{pcvstereo,
  title={Parameterized cost volume for stereo matching},
  author={Zeng, Jiaxi and Yao, Chengtang and Yu, Lidong and Wu, Yuwei and Jia, Yunde},
  booktitle={Proceedings of the IEEE/CVF International Conference on Computer Vision},
  pages={18347--18357},
  year={2023}
}

@inproceedings{igevstereo,
  title={Iterative geometry encoding volume for stereo matching},
  author={Xu, Gangwei and Wang, Xianqi and Ding, Xiaohuan and Yang, Xin},
  booktitle={Proceedings of the IEEE/CVF conference on computer vision and pattern recognition},
  pages={21919--21928},
  year={2023}
}

@inproceedings{goatstereo,
  title={Global occlusion-aware transformer for robust stereo matching},
  author={Liu, Zihua and Li, Yizhou and Okutomi, Masatoshi},
  booktitle={Proceedings of the IEEE/CVF Winter Conference on Applications of Computer Vision},
  pages={3535--3544},
  year={2024}
}

@inproceedings{selectivestereo,
  title={Selective-stereo: Adaptive frequency information selection for stereo matching},
  author={Wang, Xianqi and Xu, Gangwei and Jia, Hao and Yang, Xin},
  booktitle={Proceedings of the IEEE/CVF Conference on Computer Vision and Pattern Recognition},
  pages={19701--19710},
  year={2024}
}

@inproceedings{greatstereo,
  title={Global regulation and excitation via attention tuning for stereo matching},
  author={Li, Jiahao and Chen, Xinhong and Jiang, Zhengmin and Zhou, Qian and Li, Yung-Hui and Wang, Jianping},
  booktitle={Proceedings of the IEEE/CVF International Conference on Computer Vision},
  pages={25539--25549},
  year={2025}
}

@article{igev++,
  title={Igev++: Iterative multi-range geometry encoding volumes for stereo matching},
  author={Xu, Gangwei and Wang, Xianqi and Zhang, Zhaoxing and Cheng, Junda and Liao, Chunyuan and Yang, Xin},
  journal={IEEE Transactions on Pattern Analysis and Machine Intelligence},
  year={2025},
  publisher={IEEE}
}

@inproceedings{mgstereo,
  title={Diving into the Fusion of Monocular Priors for Generalized Stereo Matching},
  author={Yao, Chengtang and Yu, Lidong and Liu, Zhidan and Zeng, Jiaxi and Wu, Yuwei and Jia, Yunde},
  booktitle={Proceedings of the IEEE/CVF International Conference on Computer Vision},
  pages={14887--14897},
  year={2025}
}

@inproceedings{bridgedepth,
  title={Bridgedepth: Bridging monocular and stereo reasoning with latent alignment},
  author={Guan, Tongfan and Guo, Jiaxin and Wang, Chen and Liu, Yun-Hui},
  booktitle={Proceedings of the IEEE/CVF International Conference on Computer Vision},
  pages={27681--27691},
  year={2025}
}

@inproceedings{monsterstereo,
  title={Monster: Marry monodepth to stereo unleashes power},
  author={Cheng, Junda and Liu, Longliang and Xu, Gangwei and Wang, Xianqi and Zhang, Zhaoxing and Deng, Yong and Zang, Jinliang and Chen, Yurui and Cai, Zhipeng and Yang, Xin},
  booktitle={Proceedings of the Computer Vision and Pattern Recognition Conference},
  pages={6273--6282},
  year={2025}
}

@article{monster++,
  title={MonSter++: Unified Stereo Matching, Multi-view Stereo, and Real-time Stereo with Monodepth Priors},
  author={Cheng, Junda and Liao, Wenjing and Cai, Zhipeng and Liu, Longliang and Xu, Gangwei and Wang, Xianqi and Wang, Yuzhou and Yuan, Zikang and Deng, Yong and Zang, Jinliang and others},
  journal={arXiv preprint arXiv:2501.08643},
  year={2025}
}

@inproceedings{defomstereo,
  title={Defom-stereo: Depth foundation model based stereo matching},
  author={Jiang, Hualie and Lou, Zhiqiang and Ding, Laiyan and Xu, Rui and Tan, Minglang and Jiang, Wenjie and Huang, Rui},
  booktitle={Proceedings of the Computer Vision and Pattern Recognition Conference},
  pages={21857--21867},
  year={2025}
}

@inproceedings{foundationstereo,
  title={Foundationstereo: Zero-shot stereo matching},
  author={Wen, Bowen and Trepte, Matthew and Aribido, Joseph and Kautz, Jan and Gallo, Orazio and Birchfield, Stan},
  booktitle={Proceedings of the IEEE/CVF conference on computer vision and pattern recognition},
  pages={5249--5260},
  year={2025}
}

@inproceedings{stereoanywhere,
  title={Stereo anywhere: Robust zero-shot deep stereo matching even where either stereo or mono fail},
  author={Bartolomei, Luca and Tosi, Fabio and Poggi, Matteo and Mattoccia, Stefano},
  booktitle={Proceedings of the IEEE/CVF conference on computer vision and pattern recognition},
  pages={1013--1027},
  year={2025}
}

@article{promptstereo,
  title={PromptStereo: Zero-Shot Stereo Matching via Structure and Motion Prompts},
  author={Wang, Xianqi and Yang, Hao and Wang, Hangtian and Cheng, Junda and Xu, Gangwei and Lin, Min and Yang, Xin},
  journal={arXiv preprint arXiv:2603.01650},
  year={2026}
}

@article{adamw,
  title={Decoupled weight decay regularization},
  author={Loshchilov, Ilya and Hutter, Frank},
  journal={arXiv preprint arXiv:1711.05101},
  year={2017}
}

@article{transformer,
  title={Attention is all you need},
  author={Ashish, Vaswani},
  journal={Advances in neural information processing systems},
  volume={30},
  pages={I},
  year={2017}
}

@inproceedings{mobilenetv2,
  title={Mobilenetv2: Inverted residuals and linear bottlenecks},
  author={Sandler, Mark and Howard, Andrew and Zhu, Menglong and Zhmoginov, Andrey and Chen, Liang-Chieh},
  booktitle={Proceedings of the IEEE conference on computer vision and pattern recognition},
  pages={4510--4520},
  year={2018}
}

@inproceedings{onecycle,
  title={Super-convergence: Very fast training of neural networks using large learning rates},
  author={Smith, Leslie N and Topin, Nicholay},
  booktitle={Artificial intelligence and machine learning for multi-domain operations applications},
  volume={11006},
  pages={369--386},
  year={2019},
  organization={SPIE}
}

@inproceedings{raft,
  title={Raft: Recurrent all-pairs field transforms for optical flow},
  author={Teed, Zachary and Deng, Jia},
  booktitle={European conference on computer vision},
  pages={402--419},
  year={2020},
  organization={Springer}
}

@article{vit,
  title={An image is worth 16x16 words: Transformers for image recognition at scale},
  author={Dosovitskiy, Alexey and Beyer, Lucas and Kolesnikov, Alexander and Weissenborn, Dirk and Zhai, Xiaohua and Unterthiner, Thomas and Dehghani, Mostafa and Minderer, Matthias and Heigold, Georg and Gelly, Sylvain and others},
  journal={arXiv preprint arXiv:2010.11929},
  year={2020}
}

@article{deformdetr,
  title={Deformable detr: Deformable transformers for end-to-end object detection},
  author={Zhu, Xizhou and Su, Weijie and Lu, Lewei and Li, Bin and Wang, Xiaogang and Dai, Jifeng},
  journal={arXiv preprint arXiv:2010.04159},
  year={2020}
}

@article{gans,
  title={Generative adversarial networks},
  author={Goodfellow, Ian and Pouget-Abadie, Jean and Mirza, Mehdi and Xu, Bing and Warde-Farley, David and Ozair, Sherjil and Courville, Aaron and Bengio, Yoshua},
  journal={Communications of the ACM},
  volume={63},
  number={11},
  pages={139--144},
  year={2020},
  publisher={ACM New York, NY, USA}
}

@article{layernorm,
  title={Rethinking skip connection with layer normalization in transformers and resnets},
  author={Liu, Fenglin and Ren, Xuancheng and Zhang, Zhiyuan and Sun, Xu and Zou, Yuexian},
  journal={arXiv preprint arXiv:2105.07205},
  year={2021}
}

@inproceedings{gmaflow,
  title={Learning to estimate hidden motions with global motion aggregation},
  author={Jiang, Shihao and Campbell, Dylan and Lu, Yao and Li, Hongdong and Hartley, Richard},
  booktitle={Proceedings of the IEEE/CVF international conference on computer vision},
  pages={9772--9781},
  year={2021}
}

@inproceedings{mae,
  title={Masked autoencoders are scalable vision learners},
  author={He, Kaiming and Chen, Xinlei and Xie, Saining and Li, Yanghao and Doll{\'a}r, Piotr and Girshick, Ross},
  booktitle={Proceedings of the IEEE/CVF conference on computer vision and pattern recognition},
  pages={16000--16009},
  year={2022}
}

@inproceedings{deformvit,
  title={Vision transformer with deformable attention},
  author={Xia, Zhuofan and Pan, Xuran and Song, Shiji and Li, Li Erran and Huang, Gao},
  booktitle={Proceedings of the IEEE/CVF conference on computer vision and pattern recognition},
  pages={4794--4803},
  year={2022}
}

@article{bevformer,
  title={Bevformer: learning bird’s-eye-view representation from lidar-camera via spatiotemporal transformers},
  author={Li, Zhiqi and Wang, Wenhai and Li, Hongyang and Xie, Enze and Sima, Chonghao and Lu, Tong and Yu, Qiao and Dai, Jifeng},
  journal={IEEE Transactions on Pattern Analysis and Machine Intelligence},
  volume={47},
  number={3},
  pages={2020--2036},
  year={2024},
  publisher={IEEE}
}

@inproceedings{gaussianformer,
  title={Gaussianformer: Scene as gaussians for vision-based 3d semantic occupancy prediction},
  author={Huang, Yuanhui and Zheng, Wenzhao and Zhang, Yunpeng and Zhou, Jie and Lu, Jiwen},
  booktitle={European Conference on Computer Vision},
  pages={376--393},
  year={2024},
  organization={Springer}
}

@article{dinov2,
  title={Dinov2: Learning robust visual features without supervision},
  author={Oquab, Maxime and Darcet, Timoth{\'e}e and Moutakanni, Th{\'e}o and Vo, Huy and Szafraniec, Marc and Khalidov, Vasil and Fernandez, Pierre and Haziza, Daniel and Massa, Francisco and El-Nouby, Alaaeldin and others},
  journal={arXiv preprint arXiv:2304.07193},
  year={2023}
}

@article{depthanythingv2,
  title={Depth anything v2},
  author={Yang, Lihe and Kang, Bingyi and Huang, Zilong and Zhao, Zhen and Xu, Xiaogang and Feng, Jiashi and Zhao, Hengshuang},
  journal={Advances in Neural Information Processing Systems},
  volume={37},
  pages={21875--21911},
  year={2024}
}

@inproceedings{gaussianformerv2,
  title={Gaussianformer-2: Probabilistic gaussian superposition for efficient 3d occupancy prediction},
  author={Huang, Yuanhui and Thammatadatrakoon, Amonnut and Zheng, Wenzhao and Zhang, Yunpeng and Du, Dalong and Lu, Jiwen},
  booktitle={Proceedings of the computer vision and pattern recognition conference},
  pages={27477--27486},
  year={2025}
}

@inproceedings{kitti2012,
  title={Are we ready for autonomous driving? the kitti vision benchmark suite},
  author={Geiger, Andreas and Lenz, Philip and Urtasun, Raquel},
  booktitle={2012 IEEE conference on computer vision and pattern recognition},
  pages={3354--3361},
  year={2012},
  organization={IEEE}
}

@inproceedings{middlebury,
  title={High-resolution stereo datasets with subpixel-accurate ground truth},
  author={Scharstein, Daniel and Hirschm{\"u}ller, Heiko and Kitajima, York and Krathwohl, Greg and Ne{\v{s}}i{\'c}, Nera and Wang, Xi and Westling, Porter},
  booktitle={German conference on pattern recognition},
  pages={31--42},
  year={2014},
  organization={Springer}
}

@inproceedings{kitti2015,
  title={Object scene flow for autonomous vehicles},
  author={Menze, Moritz and Geiger, Andreas},
  booktitle={Proceedings of the IEEE conference on computer vision and pattern recognition},
  pages={3061--3070},
  year={2015}
}

@inproceedings{eth3d,
  title={A multi-view stereo benchmark with high-resolution images and multi-camera videos},
  author={Schops, Thomas and Schonberger, Johannes L and Galliani, Silvano and Sattler, Torsten and Schindler, Konrad and Pollefeys, Marc and Geiger, Andreas},
  booktitle={Proceedings of the IEEE conference on computer vision and pattern recognition},
  pages={3260--3269},
  year={2017}
}

@article{instereo2k,
  title={Instereo2k: a large real dataset for stereo matching in indoor scenes},
  author={Bao, Wei and Wang, Wei and Xu, Yuhua and Guo, Yulan and Hong, Siyu and Zhang, Xiaohu},
  journal={Science China Information Sciences},
  volume={63},
  number={11},
  pages={212101},
  year={2020},
  publisher={Springer}
}

@inproceedings{booster,
  title={Open challenges in deep stereo: the booster dataset},
  author={Ramirez, Pierluigi Zama and Tosi, Fabio and Poggi, Matteo and Salti, Samuele and Mattoccia, Stefano and Di Stefano, Luigi},
  booktitle={Proceedings of the IEEE/CVF Conference on Computer Vision and Pattern Recognition},
  pages={21168--21178},
  year={2022}
}

@inproceedings{sintel,
  title={A naturalistic open source movie for optical flow evaluation},
  author={Butler, Daniel J and Wulff, Jonas and Stanley, Garrett B and Black, Michael J},
  booktitle={European conference on computer vision},
  pages={611--625},
  year={2012},
  organization={Springer}
}

@inproceedings{sceneflow,
  title={A large dataset to train convolutional networks for disparity, optical flow, and scene flow estimation},
  author={Mayer, Nikolaus and Ilg, Eddy and Hausser, Philip and Fischer, Philipp and Cremers, Daniel and Dosovitskiy, Alexey and Brox, Thomas},
  booktitle={Proceedings of the IEEE conference on computer vision and pattern recognition},
  pages={4040--4048},
  year={2016}
}

@inproceedings{fallingthings,
  title={Falling things: A synthetic dataset for 3d object detection and pose estimation},
  author={Tremblay, Jonathan and To, Thang and Birchfield, Stan},
  booktitle={Proceedings of the IEEE conference on computer vision and pattern recognition workshops},
  pages={2038--2041},
  year={2018}
}

@inproceedings{hrvs,
  title={Hierarchical deep stereo matching on high-resolution images},
  author={Yang, Gengshan and Manela, Joshua and Happold, Michael and Ramanan, Deva},
  booktitle={Proceedings of the IEEE/CVF Conference on Computer Vision and Pattern Recognition},
  pages={5515--5524},
  year={2019}
}

@article{vkitti2,
  title={Virtual kitti 2},
  author={Cabon, Yohann and Murray, Naila and Humenberger, Martin},
  journal={arXiv preprint arXiv:2001.10773},
  year={2020}
}

@inproceedings{tartanair,
  title={Tartanair: A dataset to push the limits of visual slam},
  author={Wang, Wenshan and Zhu, Delong and Wang, Xiangwei and Hu, Yaoyu and Qiu, Yuheng and Wang, Chen and Hu, Yafei and Kapoor, Ashish and Scherer, Sebastian},
  booktitle={2020 IEEE/RSJ International Conference on Intelligent Robots and Systems (IROS)},
  pages={4909--4916},
  year={2020},
  organization={IEEE}
}

@inproceedings{hypersim,
  title={Hypersim: A photorealistic synthetic dataset for holistic indoor scene understanding},
  author={Roberts, Mike and Ramapuram, Jason and Ranjan, Anurag and Kumar, Atulit and Bautista, Miguel Angel and Paczan, Nathan and Webb, Russ and Susskind, Joshua M},
  booktitle={Proceedings of the IEEE/CVF international conference on computer vision},
  pages={10912--10922},
  year={2021}
}

\begin{IEEEbiography}
[{\includegraphics[width=1in,height=1.25in,clip,keepaspectratio]{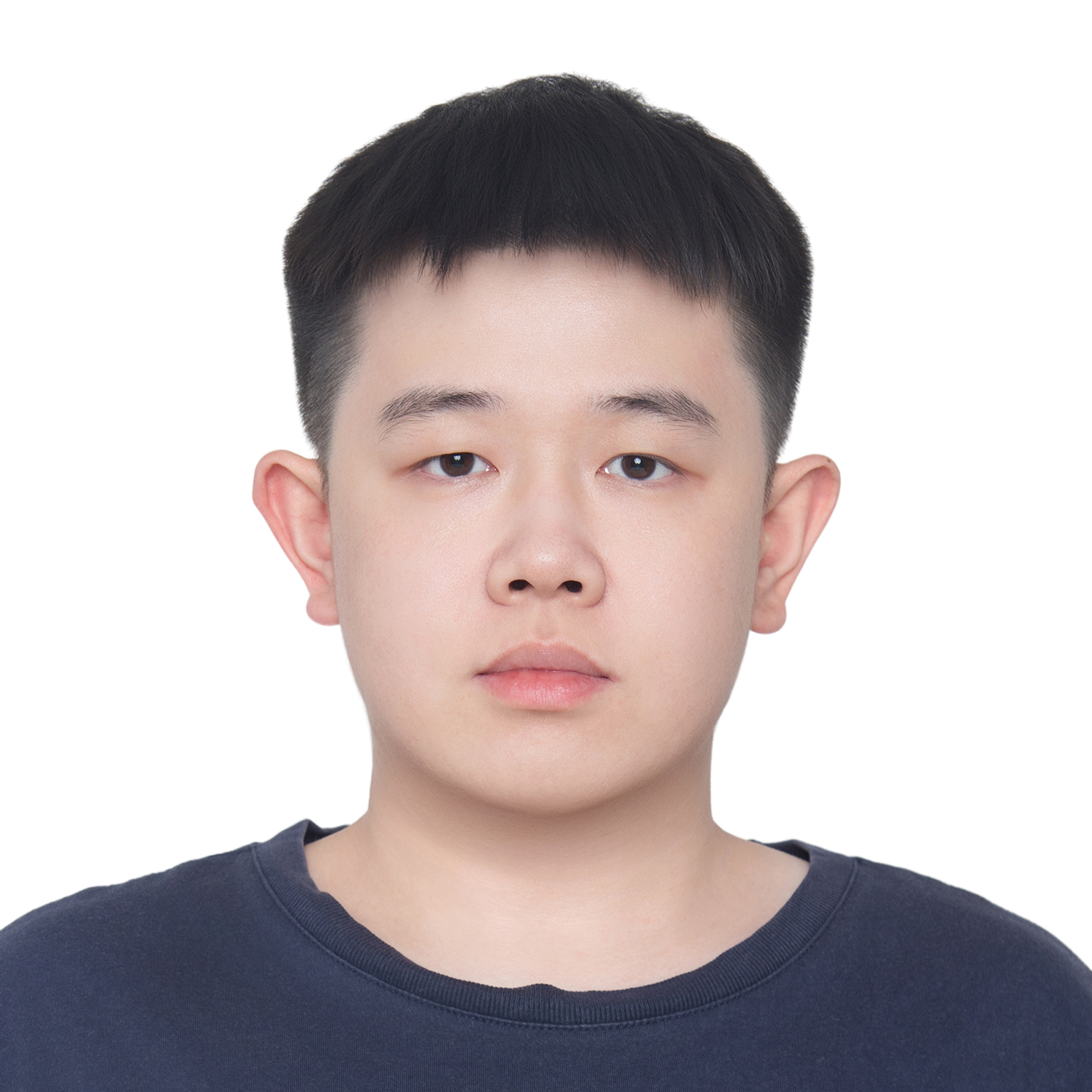}}]{Jiahao Li} is a PhD student at the Department of Computer Science at City University of Hong Kong, supervised by Prof. Jianping Wang. He received his B.S. degree in computer science and technology from Beijing Normal-Hong Kong Baptist University, Guangdong, China, in 2021, and his M.S. degree in computer science from The Chinese University of Hong Kong, Hong Kong, in 2022. His research interests include stereo matching, 3D semantic occupancy prediction, and autonomous driving perception.
\end{IEEEbiography}

\begin{IEEEbiography}[{\includegraphics[width=1in,height=1.25in,clip,keepaspectratio]{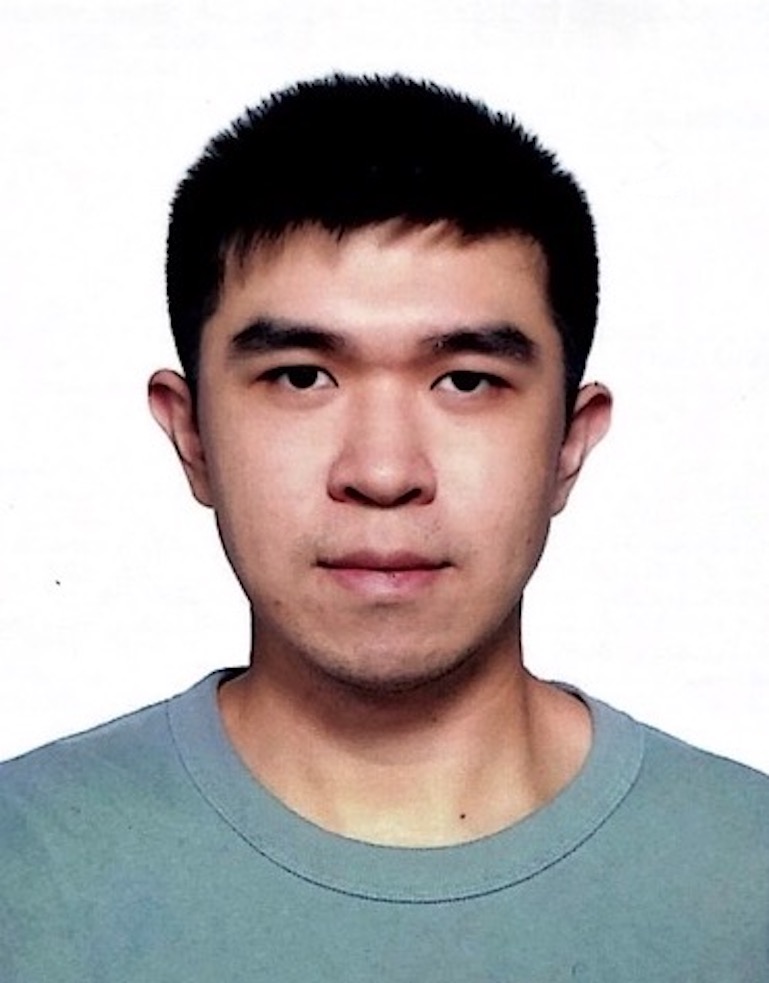}}]{Xinhong Chen} received the B.Eng. degree in software engineering from Sun Yat-Sen University, Guangdong, China, in 2018, and the Ph.D. degree in computer science from City University of Hong Kong, in 2022. He is currently a Research Assistant Professor with the Department of Computer Science at City University of Hong Kong, Hong Kong. His research interests including natural language processing, sentiment analysis, autonomous driving, and causality mining.
\end{IEEEbiography}

\begin{IEEEbiography}[{\includegraphics[width=1in,height=1.25in,clip,keepaspectratio]{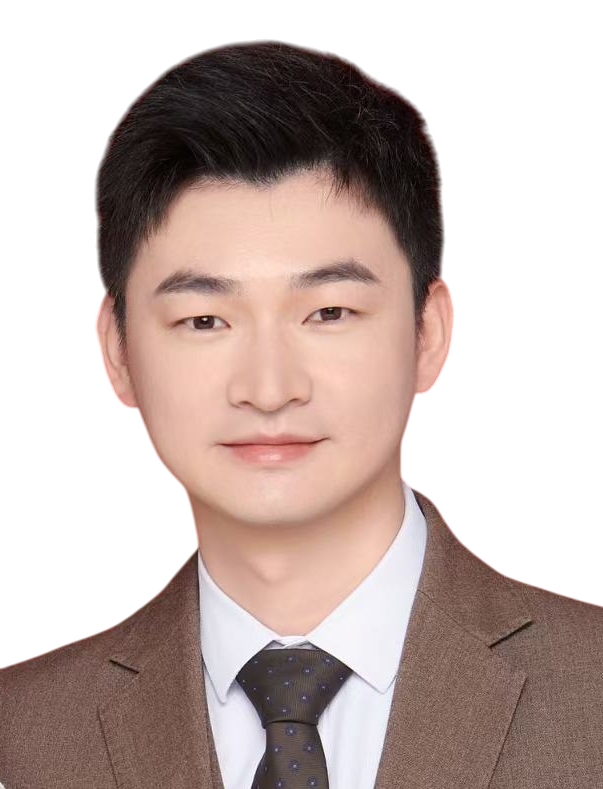}}]{Zhengmin Jiang} received his MPhil degree in Automotive Engineering from Chang’an University, Xi’an, China, in 2018. From 2018 to 2020, he worked as an autonomous driving software engineer at the SAIC Motor Technical Centre. He subsequently earned his Ph.D. degree in Pattern Recognition and Intelligent Systems from the University of Chinese Academy of Sciences (UCAS). Currently, he is a Postdoctoral Fellow in the Department of Computer Science, City University of Hong Kong (CityUHK). His research interests include autonomous vehicle safety, generative artificial intelligence, and reinforcement learning.
\end{IEEEbiography}

\begin{IEEEbiography}
[{\includegraphics[width=1in,height=1.25in,clip,keepaspectratio]{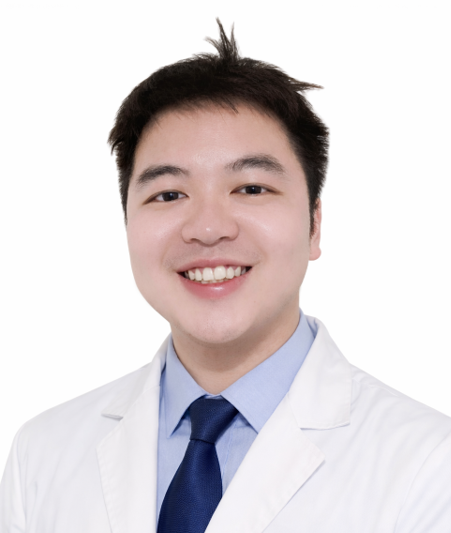}}]{Cheng Huang} received the B.S. degree in electronic engineering from the University of Electronic Science and Technology of China, Chengdu, China, in 2020, and the M.S. degree in information engineering from The Chinese University of Hong Kong, Hong Kong, in 2022. He received the Ph.D. degree in computer science from Southern Methodist University (SMU), Dallas, TX, USA, in 2025. His research interests include medical artificial intelligence, multimodal learning, and data-centric modeling for ophthalmic diseases. His work focuses on developing clinically interpretable AI systems that integrate imaging and clinical data to support diagnosis and disease progression analysis, particularly in glaucoma.
\end{IEEEbiography}

\begin{IEEEbiography}
[{\includegraphics[width=1in,height=1.25in,clip,keepaspectratio]{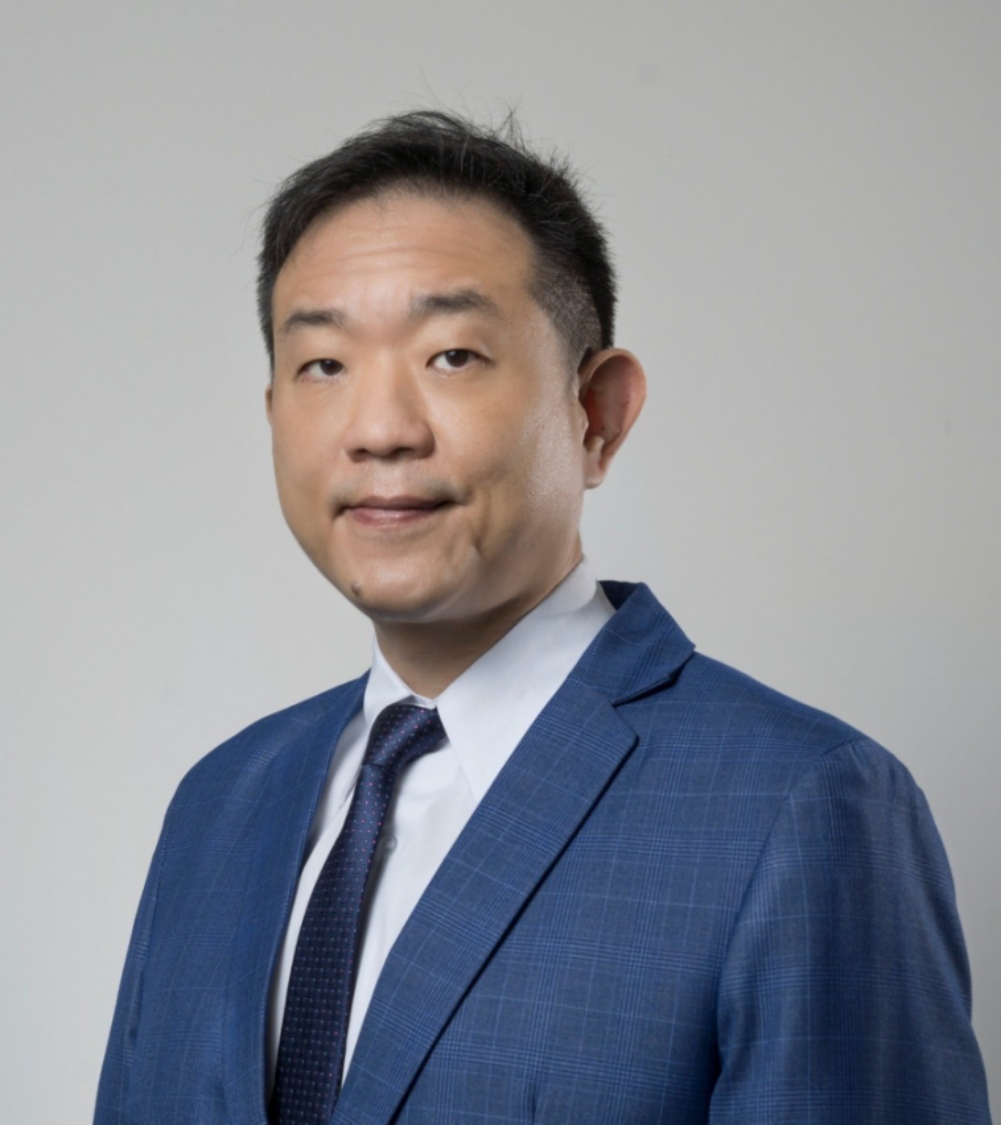}}]{Yung-Hui Li} (Member, IEEE) received the B.S. degree from National Taiwan University, Taipei City, in 1995, the M.S. degree from the University of Pennsylvania, Philadelphia, PA, USA, in 1998, and the Ph.D. degree from the School of Computer Science, Carnegie Mellon University, Pittsburgh, PA, in 2010. He is currently the founding Director of the AI Research Center, Hon Hai Research Institute, Taipei City, Taiwan, a pivotal R\&D hub driving Foxconn’s 3+3 transformation strategy. Prior to his association with Foxconn, he held a tenured faculty position with National Central University, Taoyuan, Taiwan. His research team has made significant strides in the field of autonomous driving, garnering recognition through several gold and silver medals at renowned events like the Geneva International Exhibition of Inventions, Pittsburgh International Invention Show (INPEX), and Silicon Valley International Invention Festival (SVIIF). His current research interests include AI, deep learning, computer vision, multimedia, multimodal foundation models, autonomous driving, and biometric recognition. He actively contributes to the international academic community, having served in various capacities, including industry chair, publicity chair, and technical program committee member for numerous conferences.
\end{IEEEbiography}

\begin{IEEEbiography}
[{\includegraphics[width=1in,height=1.25in,clip,keepaspectratio]{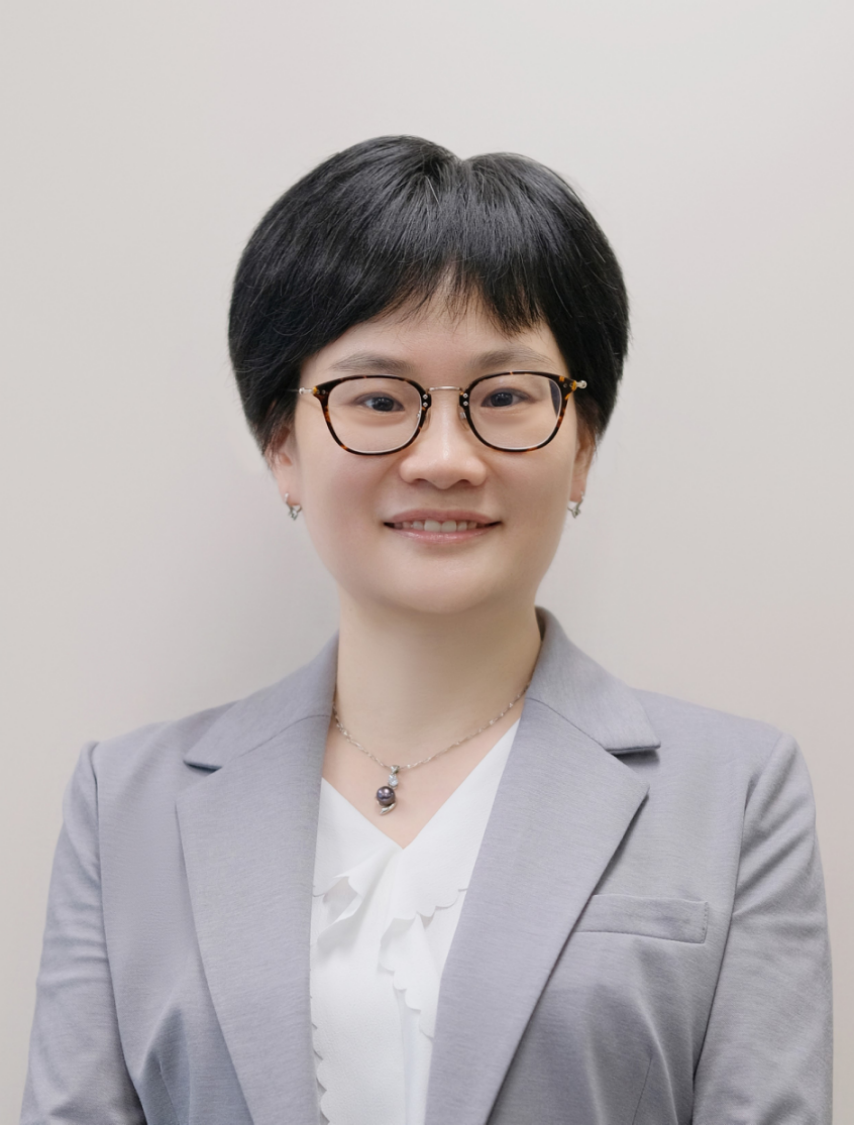}}]{Jianping Wang} (Fellow, IEEE) received the B.S. and M.S. degrees in computer science from Nankai University, Tianjin, China, in 1996 and 1999, respectively, and the Ph.D. degree in computer science from The University of Texas at Dallas, in 2003. She is currently a Chair Professor with the Department of Computer Science, City University of Hong Kong. Her research interests include security, autonomous driving, cloud computing, edge computing, and machine learning.
\end{IEEEbiography}


 




\vfill

\end{document}